%% file: main.tex
\documentclass[twoside]{article}

\usepackage[preprint]{aistats2026}
\usepackage[round]{natbib}

\usepackage[utf8]{inputenc} 

\usepackage{xurl}
\usepackage[T1]{fontenc}    
\usepackage{url}            
\usepackage{booktabs}       
\usepackage{amsfonts}       
\usepackage{nicefrac}       
\usepackage{microtype}      
\usepackage{xcolor}         
\usepackage{amsmath}
\usepackage{physics}
\usepackage{bm}
\usepackage{subcaption}
\usepackage{graphicx}
\usepackage{float}
\usepackage{tikz}
\usepackage{pgfplots}
\usepackage{makecell}
\usepackage{booktabs}
\usepackage[urlbordercolor={1 1 1}]{hyperref}       
\usepackage{algorithm}
\usepackage{algpseudocode}
\usepackage{enumitem}

\usepackage{xcolor}
\usepackage{cleveref}

\newcommand{\marker}[1]{\color{#1}}

\definecolor{custom_green}{HTML}{008000} 
\definecolor{custom_yellow}{HTML}{FEDD00} 

\begin{document}

%

%
\makeatletter
\AtBeginDocument{
  \gdef\@runningtitle{A}
  \gdef\@runningauthor{A}
}
\makeatother

\twocolumn[
\aistatstitle{Warm-Starting Iterative Gaussian Processes for\\Faster Sequential Inference}
\aistatsauthor{ Alan Yufei Dong \And Jihao Andreas Lin \And  Jos\'e Miguel Hern\'andez-Lobato }
\aistatsaddress{Department of Engineering, University of Cambridge}]

\begin{abstract}
Efficient Gaussian process (GP) inference is critical for sequential decision-making tasks such as active learning, online prediction, and Bayesian optimization. Iterative approaches of approximating the GP posterior using solvers like conjugate gradients, stochastic gradient descent, or alternating projections avoid cubic costs, but often require many iterations to converge, limiting their efficacy when the posterior is updated frequently with new data. To address this, we introduce three warm-start strategies that exploit solutions of smaller linear systems to substantially speed-up convergence when updating the posterior with new data. Our methods are supported by theoretical analysis showing reduced initialization error in reproducing kernel Hilbert space (RKHS) distance, and by empirical results on regression benchmarks and Bayesian optimization tasks. Across solvers, warm-starting achieves speed-ups of up to 19× when solving to tolerance, and produces more accurate posterior estimates under fixed compute budgets, directly improving optimization performance. These results establish warm-starting as a simple, effective, and broadly applicable tool for scaling Gaussian processes in sequential settings.
\end{abstract}

\section{Introduction}
\label{introduction}

\input{introduction}

\section{Sampling from the Gaussian Process Posterior}
\label{background}

\input{Section_2/background}

\paragraph{Pathwise Conditioning}

\input{Section_2/pathwise}

\paragraph{Using Iterative Solvers}

\input{Section_2/iterativesolvers}

\section{Linear Systems During Sequential Inference}
\label{section:linear systems}

\input{Section_2/newdata}

\subsection{Initial Distance of Weights from Their Solution}
\label{subsec:initialdistance}
 \citet{lin2024warm} uses the reproducing kernel Hilbert space (RKHS) distance to provide theoretical arguments for warm-starting during hyperparameter tuning with a fixed dataset. We make a similar argument to support our methods for warm-starting the solve for the extended linear system \eqref{eq:extended_system_main_text} that arises when updating the GP posterior with additional data points.
 
The squared RKHS distance of the weights from their exact solutions is defined as,
\vspace{10pt}
\begin{equation}
    d^2 = \left\|\bm{v}-\bm{v}^*\right\|^2_{\vb{H}}=\left(\bm{v}-\bm{v}^*\right)^\top\vb{H}\left(\bm{v}-\bm{v}^*\right).
\end{equation}

Let $d_\text{cold}^2$ denote the initial squared RKHS distance of the weights when solving from scratch (cold-start), and let $d_\text{na\"ive}^2$ denote the distance under the naive warm-start method \eqref{eq:method1}. The difference between these two squared distances is

\begin{equation}
    d_\text{cold}^2-d_\text{na\"ive}^2=\bm{b}_1^\top\vb{H}_{11}^{-1}\bm{b}_1\geq0,
\end{equation}

\vspace{5pt}

where $\bm{b}_1$ is the first block of the right-hand side vector from the extended linear system \eqref{eq:extended_system_main_text}.

The final inequality comes from the fact that $\vb{H}_{11}^{-1}$ is positive definite (as a GP covariance matrix is always positive definite), with equality only when $\bm{b}_1=\bm{0}$. Hence, warm-starting with the na\"ive method strictly reduces the initial distance to the solution. In general, this reduction will also be a greater proportion of the cold-start distance if $n_1$ is larger relative to $n_2$, since the $\bm{b}_1^\top\vb{H}_{11}^{-1}\bm{b}_1$ term is more dominant compared to the other terms (see \eqref{eq:cold_rkhs} in \Cref{appendix:derivation}).

Let $d_\text{line-search}^2$ denote the initial squared RKHS distance of the weights under the residual line search method \eqref{eq:method2}, and let $d_\text{marginal-solve}^2$ denote the distance under the marginal system solve method \eqref{eq:method3}. Performing an additional residual line search or marginal system solve further reduces the initial distance:

\begin{equation}
 d^2_\text{na\"ive}-d^2_\text{line-search}= \frac{(\bm{r}^\top \bm{r})^2}{\bm{r}^\top \vb{H}_{22}\bm{r}} > 0
\end{equation}

\begin{equation}
   d^2_\text{na\"ive}-d^2_\text{marginal-solve} = \bm{r}^\top \vb{H}_{22}^{-1} \bm{r} >0
\end{equation}

Furthermore, the marginal system solve method is a more accurate initialization than the reduced line search method, as
\begin{equation}
\bm{r}^\top \vb{H}_{22}^{-1} \bm{r}\geq \frac{(\bm{r}^\top \bm{r})^2}{\bm{r}^\top \vb{H}_{22}\bm{r}}.
\end{equation}

\vspace{5pt}

The complete derivations are provided in \Cref{appendix:derivation}, where we additionally explain that reducing the RKHS distance of the weights from the exact solution is equivalent to optimizing the quadratic objective, and equivalent to reducing the residual of the linear system that is used as a stopping criterion for the solvers. This means that our methods exploit the intrinsic operating principles behind iterative linear solvers.

\section{Experiments}
\label{section:experiments}

We performed experiments in two machine learning tasks to study the effectiveness of warm-starting sequential posteriors in two different settings:

\begin{enumerate}
    \item A regression task using real-world datasets to show that our methods speed-up the computation of posterior samples to the same level of accuracy.
    \item A Bayesian optimization task with parallel Thompson sampling to show that warm-starting achieves more accurate posteriors and better performance under the same limited compute budget.
\end{enumerate}

\subsection{Regression Experiments}
\label{section:regression}
\input{Section_3/regression_experiment}



\paragraph{Discussion}
\Cref{fig:initial_distance_mean} shows the differences in initial distances and \Cref{fig:solver_iterations_mean} shows the differences in number of solver iterations for the posterior mean system. \Cref{fig:initial_distance_mean} shows that the na\"ive method consistently places the initial weights approximately 70\% closer to their final solution on average compared to initializing the weights at 0, for all datasets. 
Incorporating initial values for the new weights yields on average a further 4\% reduction with a residual line search method, and 7\% with a marginal system solve. As shown in \Cref{fig:solver_iterations_mean}, a smaller initial distance generally corresponds to fewer iterations required by all solvers, although for CG this trend is not observed with the residual line search method for some datasets. The degree of reduction varies between datasets and the solver used.

On average, across all datasets and both the posterior mean and sample linear systems, the na\"ive method reduces the solver iterations by approximately 30\% for CG, 34\% for SGD and 75\% for AP compared to the cold-start baseline. This is equivalent to speed-ups of 1.4$\times$, 1.5$\times$ and 4.0$\times$ for CG, SGD and AP respectively. The reduction is particularly significant for AP -- up to 94.6\% for some datasets, equivalent to a 19$\times$ speed-up.

Method 2 -- performing a residual line search -- achieves average reductions in the number of solver iterations of 25\%, 37\% and 76\%, or equivalently speed-ups of 1.3$\times$, 1.6$\times$ and 4.2$\times$.

Finally, Method 3 -- solving the marginal system -- achieves average reductions in the number of solver iterations of 33\%, 41\% and 76\%, or equivalently speed-ups of 1.5$\times$, 1.7$\times$ and 4.2$\times$.

While the warm-start methods in general are most effective with AP, the two methods which further reduce the initial distance by specifying values for the new weights $\bm{v}_2$ exhibit the most significant improvements over the na\"ive warm-start method with SGD. For CG, the relative iterations for the residual line search method are occasionally not an improvement over the na\"ive warm-start method. This is because convergence for CG depends not only on the initial residual’s magnitude but also on how its direction aligns with the eigenvectors of $\vb{H}$ \citep{liesen2004convergence}.

\input{Section_3/bayes_opt_results}

These results on real-world datasets support the theoretical case for warm-starting discussed previously. However, we note that these speed-ups are specific to the particular ratio of 1000 to 100 data points; referring back to \Cref{subsec:initialdistance}, warm-starting on average achieves greater speed-ups if the number of new data points is smaller in proportion to the existing dataset, and less significant speed-ups if the proportion is larger. The regression experiment results for the posterior sample system are very similar to that which we presented for the posterior mean system; we include both together in \Cref{appendix:regression_results}.

\subsection{Parallel Thompson Sampling Experiments}

\input{Section_3/bayes_opt_experiment}

\paragraph{Discussion}
Shown in \Cref{fig:small budget} is the highest value found on the objective function and the normalized final residual of the posterior mean solve $\lVert \vb{H} \bm{v} - \bm{y} \rVert/\lVert \bm{y} \rVert$, after each batch of point acquisitions. Under a limited compute budget with a constant cap on solver iterations (5, 120 and 30 iterations respectively for CG, SGD and AP), using more accurate estimates of the exact solution to warm-start the linear solves yields smaller final residuals. Solver progress accumulates over multiple solvers instead of resetting after each solve. This gives more accurate posterior predictions which improve the ability of the optimizer to find the maximum of the objective.

This means that in cases where a typically cold-started solve gives inaccurate GP posteriors, our methods are effective at maintaining accurate posteriors. Even when previous linear solves have not fully converged, accurately initializing the representer weights of the subsequent linear solves nonetheless places the subsequent solves closer to their final solution, resulting in smaller final residuals.

With SGD, the maximum function value found on the objective function as points are acquired clearly shows the relative performance of the three proposed methods and the cold-start baseline, which reflects the theoretical analysis of the initial distance in \Cref{subsec:initialdistance}. For CG and AP, we see a scenario where the na\"ive method of simply using the previous weights is accurate enough for the given compute budget such that the additional reductions in distance of the improved warm-start methods do not significantly improve the performance further. Measuring the final maximum value found on the objective function, our methods achieve performance improvements of up to 13\% for CG, 46\% for SGD, and 10\% for AP.

In summary, the introduced methods are successful at improving the accuracy of GP posterior predictions during sequential inference. While they offer different levels of accuracy, each is demonstrably useful and can be applied effectively in different scenarios depending on dataset size, computational budget, and other practical constraints. Full results for each lengthscale are listed individually in \Cref{appendix:bayes_opt_results}, together with the final residuals of the posterior sample system.

\section{Conclusion}
\input{conclusion}

\bibliographystyle{plainnat}
\bibliography{references}



\newpage
\onecolumn
\appendix
\crefalias{section}{appendix}

\section*{\MakeUppercase{Appendices}}

\section{Initial Distance to Solution -- Full Derivation}\label{appendix:derivation}
\input{Appendices/derivation}

\newpage

\section{Implementation Details}\label{appendix:implementation}
\input{Appendices/implementation}

\newpage

\section{Regression Experiment -- Full Results}\label{appendix:regression_results}
\input{Appendices/regression_results}

\newpage

\section{Parallel Thompson Experiment -- Full Results}\label{appendix:bayes_opt_results}
This appendix contains the full results for the parallel Thompson experiment, including the results for each individual lengthscale.
\input{Appendices/bayes_opt_results}

\newpage

\section{Finding the Maximizing Inputs of Posterior Samples}\label{appendix:parallelThompson}
\input{Appendices/Thompson_method}



\end{document}

%% file: introduction.tex
Gaussian processes (GPs) are a powerful class of non-parametric models which have been adopted for machine learning tasks such as regression, classification, and Bayesian optimization \citep{rasmusen2006gaussian}. Despite their flexibility and ability to provide uncertainty quantification, the primary limitation of GPs lies in their poor scalability with dataset size due to the inversion of the covariance matrix which is required to compute the GP posterior. There are largely two approaches used in the literature to tackle the challenge of scalability: sparse methods and iterative methods. Sparse methods \citep{quinonero2005unifying} replace the full kernel matrix with low-rank approximations, which enables faster inversion but at the cost of reduced accuracy that can lead to poor fit on highly complex data \citep{sampling_from_gp_posteriors}. Where sparse methods reduce complexity by approximating the GP posterior itself, iterative methods \citep{gibbs1997efficient} maintain the exact posterior but use iterative linear system solvers to balance accuracy and compute time of the numerical solution.

In this paper, we focus on iterative GPs in sequential settings, where scalability is particularly critical because the incremental addition of data points necessitates continual model updates. Examples of such settings include active learning \citep{kapoor2007active}, online learning \citep{maiworm2021online} and Bayesian optimization \citep{kandasamy2018parallelised}.
To obtain updated predictions in such scenarios, the GP posterior must be recomputed by solving an increasingly larger linear system that incorporates the newly added data points.

Iterative solvers iteratively refining the estimate of the solution from an initial set of values by minimizing a corresponding quadratic objective. By stopping the solvers once a certain tolerance is reached and before full convergence, a sufficiently accurate solution can be obtained without the cubic order costs of exact algorithms like Gaussian elimination or matrix decompositions. The na\"ive approach of using iterative solvers is to solve each new larger linear system independently, initializing all values to zero and discarding any information from the previous computations.
In contrast, we propose three initialization strategies for the extended system that exploit the solution of the previous, smaller system contained within it.

The general approach of using previous values as the initial point of some new optimization task is often referred to as \textit{warm-starting}. Warm-start techniques have been effectively applied to many mathematical optimization and machine learning problems in the literature: using the solution of a linear program to warm-start the solution of a linear program with slightly different data \citep{linear_program_warm_start}; warm-starting parameter training of a neural network with additional training data \citep{neural_network_warm_start}; and warm-starting marginal log-likelihood optimization during hyperparameter tuning of a Gaussian process \citep{improving_linear_system_solvers}.

However, a common characteristic across these examples is the fixed dimensionality of the optimization problem -- that is, the number of parameters or weights being optimized or solved for is constant throughout the incrementally evolving optimization problem. We instead focus on sequential problems with Gaussian processes, where the dimensionality of the problem grows as the number of data points increases. This presents the twofold challenge of determining effective initializations for both the original weights and the newly introduced weights, given that simply reusing the previous solution does not fully populate the larger set of weights. We show that warm-starting in this scenario can likewise accelerate convergence significantly for linear solvers such as conjugate gradients, alternating projections, and stochastic gradient descent.

We demonstrate the following:
\begin{itemize}
    \item When iteratively solving the linear system until the final residual has converged to a given tolerance, our warm-start techniques increase the compute speed by up to 2.2$\times$ for SGD, 1.8$\times$ for CG and 19.2$\times$ for AP. This is achieved by reducing the number of iterations required by iterative solvers, which reduces the cost of computing the GP posterior.
    \item When operating with a fixed maximum number of solver iterations in a parallel Thompson experiment like \citet{sampling_from_gp_posteriors}, our warm-start methods result in a smaller final residual. This results in more accurate GP posterior samples under the same time constraints, improving performance by up to 13\% for CG, 46\% for SGD and 10\% for AP.
\end{itemize}

In total, we introduce three warm-start initializations, supported by theoretical analysis and empirical results, that leverage the solution of the preceding linear solve to speed-up solver convergence.

\vfill

%% file: Section_2/background.tex
Gaussian processes model an unknown function $f(\bm{x})$ by placing a random distribution over the function, such that any finite selection of points on the function form a multivariate Gaussian distribution, defined by a mean function $m(\bm{x})= \mathbb{E}[f(\bm{x})]$ and kernel function $k(\bm{x},\bm{x}')=\mathrm{Cov}(f(\bm{x}),f(\bm{x}'))$.

An observation $y$ from the GP corresponding to inputs $\vb*{x}$ can be mathematically expressed as

\begin{equation}
\begin{gathered}
y = f(\vb*{x}) + \epsilon, \quad \epsilon \sim \mathcal{N}(0, \sigma_n^2),
 \\ f(\vb*{x}) \sim \mathcal{GP}(m(\vb*{x}), k(\vb*{x}, \vb*{x}')),
\label{eq:GP}
\end{gathered}
\end{equation}

where $\epsilon$ represents Gaussian noise with variance $\sigma_n^2$

Prior beliefs about the function are expressed probabilistically as a \textit{prior} distribution $f \sim \mathcal{GP}(m, k)$ over the function. After conditioning on an observed training dataset $\mathcal{D} = (\vb{X},\bm{y}) = \{(\vb*{x}_i, y_i)|i=1,\cdots,n\}$, the \textit{posterior} distribution $f|\vb*{y} \sim \mathcal{GP}(m_{f|\vb*{y}}, k_{f|\vb*{y}})$ can be constructed to make predictions:

\begin{equation}
    m_{f|\vb*{y}}(\cdot)=\vb{K}(\cdot,\vb{X})\;\vb{H}^{-1}\vb*{y},
    \label{eq:posterior_mean}
\end{equation}
\begin{equation}
    k_{f|\vb*{y}}(\cdot,\cdot')=k(\cdot,\cdot')-\vb{K}(\cdot,\vb{X})\;\vb{H}^{-1}\vb{K}(\vb{X},\cdot').
    \label{eq:posterior_variance}
\end{equation}

In \eqref{eq:posterior_mean} and \eqref{eq:posterior_variance},

\begin{equation}
    \vb{H} := \mathrm{Cov}(\vb*{y}, \vb*{y}') = \vb{K}(\vb{X},\vb{X})+\sigma_n^2\vb{I} \in \mathbb{R}^{n\times n}
\end{equation}

is referred to as the covariance matrix,

\begin{equation}
    \vb{K}(\vb{X},\vb{X}):=[k(\bm{x}_i,\bm{x}_j)]_{i,j=1}^n\in \mathbb{R}^{n\times n}
\end{equation}
is the kernel matrix, and 

\begin{equation}
    \vb{K}(\cdot,\vb{X}):={[k(\cdot,\bm{x}_j)]_{j=1}^n}^\top\in\mathbb{R}^{n\times 1},
\end{equation}
\begin{equation}
    \vb{K}(\vb{X},\cdot'):=[k(\bm{x}_i,\cdot')]_{i=1}^n\in\mathbb{R}^{1\times n}
\end{equation}

denote vectors of pairwise kernel evaluations.

%% file: Section_2/pathwise.tex
\citet{wilson2021pathwise} introduced pathwise conditioning, an efficient method for obtaining a sample from the GP posterior distribution directly, eliminating the need to first compute the posterior mean and covariance. The posterior sample is expressed as the random function:
\begin{equation}
\begin{gathered}
(f|\vb*{y})(\cdot) = f(\cdot) \:+ \vb{K}(\cdot,\vb{X})\;\vb{H}^{-1}(\vb*{y}-(f(\vb{X} )+\vb*{\epsilon})).
\label{eq:pathwise}
\end{gathered}
\end{equation}

The posterior sample is composed of $f$, a sample from the GP prior, and a correction term which turns the prior sample into a posterior sample with the information from the data points. Here, $\vb*{\epsilon} \sim \mathcal{N}(\vb{0}, \sigma_n^2 \mathbf{I})$ is a random noise sample of the same dimension as $\bm{y}$.

As shown in \eqref{eq:pathwise}, computing a posterior sample requires solving the linear system $\vb{H}^{-1}(\vb*{y}-(f(\vb{X} )+\vb*{\epsilon}))$.
Following \citep{sampling_from_gp_posteriors}, we split this into two terms for computation: $\vb{H}^{-1}\vb*{y}$, which gives the posterior mean, common to all posterior samples, and $\vb{H}^{-1}(f(\vb{X} )+\vb*{\epsilon})$, which gives the uncertainty reduction term computed individually for each sample since $f$ and $\vb*{\epsilon}$ will be different for each posterior sample. Henceforth, we refer to the former linear system as the \textit{posterior mean system}, and the latter as the \textit{posterior sample system}.

%% file: Section_2/iterativesolvers.tex
Computing a matrix inversion exactly via Gaussian elimination or matrix decompositions scales as $O(n^3)$ -- cubic with the size of the linear system -- making it prohibitively expensive for large-scale sequential inference, where each addition of new data requires a full cubic order solve to update the GP posterior.

One approach to compute a general matrix inversion $\vb*{v} = \vb{H}^{-1}\vb*{b}$ which cannot be handled by matrix decomposition is to use iterative solvers, such as conjugate gradients (CG) \citep{osborne2007gaussian}, stochastic gradient descent (SGD) \citep{sampling_from_gp_posteriors,shalev2013stochastic} or alternating projections (AP) \citep{alternating_projection}. This is done by minimizing the quadratic objective:

\begin{equation}
	J(\vb*{v})= {\frac{1}{2}\vb*{v}^\top \vb{H}\vb*{v}-\vb*{v}^\top \vb*{b}}
    \label{eq:quadraticobjective}
\end{equation}

until the relative residual norm $\lVert \vb{H} \vb*{v} - \vb*{b} \rVert/\|\vb*{b}\|$ is within a specified tolerance. The value of the weights $\bm{v}$ minimizing the objective in \eqref{eq:quadraticobjective} are denoted $\vb*{v}^*$ -- the exact solution of the linear system.

The chosen tolerance determines the trade-off between computation time and accuracy,
with a smaller tolerance yielding greater accuracy but longer compute times. Another way of
stopping the linear solver is to specify a maximum number of solver iterations, which we also refer to as the compute budget. 

%% file: Section_2/newdata.tex
Assume the solution $\vb*{u}_1$ to a linear system with $n_1$ data points has been obtained for the current GP posterior:

\begin{equation}
    \vb{H}_{11}\vb*{u}_1 = \vb*{b}_1
\end{equation}

In sequential settings, the GP posterior must be continually updated as new data arrive incrementally. The addition of $n_2$ additional data points necessitates the solve of an extended joint linear system to update the posterior:

\begin{equation}
    \vb{H}\vb*{v}=\left[\begin{matrix}\vb{H}_{11}&\vb{H}_{12}\\\vb{H}_{12}^\top&\vb{H}_{22}\\\end{matrix}\right]\begin{bmatrix}
\bm{v}_1 \\
\bm{v}_2
\end{bmatrix}
=
\begin{bmatrix}
\vb*{b}_1 \\
\vb*{b}_2
\end{bmatrix}
= \bm{b}.
\end{equation}

Typically, iterative solvers are initialized with all-zero values:
\begin{equation}
\label{eq:cold_start}
\vb*{v} = \begin{bmatrix}
\bm{v}_1 \\
\bm{v}_2
\end{bmatrix} \gets \begin{bmatrix}
\bm{0} \\
\bm{0}
\end{bmatrix},
\end{equation}
a \textit{cold-start} which assumes no prior information about the solution, leading to slow convergence.
For sequential inference tasks, we propose that iterative solvers can be made to converge faster by using previous solutions to \textit{warm-start} the subsequent linear solves at an initial location that is closer to the final solution, which we measure with the reproducing kernel Hilbert space (RKHS) norm. This follows \citet{lin2024warm}, where a similar argument is made for warm-starting MLL optimization for hyperparameter tuning of a GP, a problem which similarly concerns linear solves with the covariance matrix $\vb{H}$.

In that context, however, the set of data points $(\vb{X}, \vb*{y})$ remains fixed, but the hyperparameters which define the kernel,  and by extension the matrix $\vb{H}$, change slightly in each iteration of MLL optimization.
\citet{lin2024warm} uses the RKHS distance of the initial location from the final solution to show that their new estimator for the MLL gradient reduces the initial distance and thus allows linear solvers to converge faster.

Our investigation focuses on the alternative setup where new data points continuously extend the training dataset. Assuming that we have access to $\vb*{u}_1$, the solution of the original linear system, also known as the representer weights in the context of kernel methods and GPs \citep{scholkopf2001generalized}, we propose and investigate three initialization methods that can be used to warm-start the new linear solve:

\begin{enumerate}[label=\textbf{Method \arabic*:}, leftmargin=0pt, labelwidth=!, align=left, itemindent=0pt]
\item\textbf{Na\"ive warm-start}

For this method, the weights of the original data points $\bm{v}_1$ are initialized at their previous solution, while the weights of the new data points $\bm{v}_2$ are simply initialized at the origin:

\begin{equation}
\label{eq:method1}
\vb*{v} = \begin{bmatrix}
\bm{v}_1 \\
\bm{v}_2
\end{bmatrix} \gets \begin{bmatrix}
\bm{u}_1 \\
\bm{0}
\end{bmatrix}.
\end{equation}

This approach preserves progress from the previous solve while assigning a neutral starting point to the new weights.

\clearpage

\item\textbf{Residual line search} 

The na\"ive warm-start initialization leaves an initial residual of 

\begin{equation}
    \bm{r} =\begin{bmatrix}\bm{b}_1\\\bm{b}_2\end{bmatrix}-\begin{bmatrix}\vb{H}_{11} & \vb{H}_{12}\\ \vb{H}_{12}^\top & \vb{H}_{22}\end{bmatrix}\begin{bmatrix}\bm{u}_1\\\bm{0}\end{bmatrix}
=\bm{b}_2-\vb{H}_{12}^\top \bm{u}_1.
\end{equation}

To further reduce the initialization error and optimize the quadratic objective, we initialize the weights of the new data points $\bm{v}_2$ with a line search in the direction of the residual, i.e., $\bm{v}_2=\alpha \bm{r}$. The value for $\alpha$ which minimizes the objective is

\begin{equation}
\alpha^*=\frac{\bm{r}^\top\bm{r}}{\bm{r}^\top\vb{H}_{22}\bm{r}},
\end{equation}

which we derive in \Cref{appendix:derivation}. This provides an informed starting point for the new weights which reduces the initialization error. Hence, we get

\begin{equation}
\label{eq:method2}
\vb*{v} = \begin{bmatrix}
\bm{v}_1 \\
\bm{v}_2
\end{bmatrix} \gets \begin{bmatrix}
\bm{u}_1 \\
\frac{\bm{r}^\top\bm{r}}{\bm{r}^\top\vb{H}_{22}\bm{r}}\bm{r}
\end{bmatrix}.
\end{equation}

\item\textbf{Marginal system solve}

Rather than restricting the new weights $\bm{v}_2$ to a scalar multiple of $\bm{r}$, we initialize $\bm{v}_2$ with the value which minimizes the quadratic objective given the fixed initial value for $\bm{v}_1$ at $\bm{u}_1$. The value which achieves this is the solution to the marginal system $\vb{H}_{22}\bm{v}_2=\bm{r}$, which we derive in \Cref{appendix:derivation}. Hence, $\bm{v}_2=\vb{H}_{22}^{-1}\bm{r}$ is the optimal starting point for the new weights derived from knowing the solution of the previous linear system.

\begin{equation}
\label{eq:method3}
\vb*{v} = \begin{bmatrix}
\bm{v}_1 \\
\bm{v}_2
\end{bmatrix} \gets \begin{bmatrix}
\bm{u}_1 \\
\vb{H}^{-1}_{22}\bm{r}
\end{bmatrix}
\end{equation}
\end{enumerate}

The proposed initialization strategies form a hierarchy of increasing complexity -- $O(1)$, $O(n_2^2)$, and $O(n_2^3)$ -- that achieve progressively closer estimates of the exact solution. In large-scale settings where the size of the incremental data $n_2$ is much smaller than that of the existing dataset $n_1$, even the most expensive initialization can be cheaper than performing additional linear solver iterations to achieve the same level of convergence.

%% file: Section_3/regression_experiment.tex
The RKHS distance to the exact solution of the extended linear system was shown analytically to be reduced by our initialization methods. We performed experiments using regression datasets from the popular UC Irvine machine learning repository \citep{asuncion2007uci} to demonstrate the empirical effectiveness of warm-starting iterative linear solvers for GP regression. We used the following datasets normalized with zero mean and unit variance: \texttt{3droad} (\texttt{3drd}), \texttt{pol}, \texttt{protein} (\texttt{prot}), \texttt{bike}, and \texttt{buzz}. These datasets cover a range of sizes, dimensionalities, and tuned hyperparameters (using the Matérn-$\frac{3}{2}$ kernel), listed in \Cref{tab:datasets_main_text}.

{\setlength{\tabcolsep}{3pt}
\begin{table}[htbp]
\centering
\caption{Dimensionality, size and tuned hyperparameters of datasets used in experiments. The hyperparameters listed are the average across the 10 trials of the experiment, and are found for each subset of data by optimizing the marginal log-likelihood.}
\vspace{0.5em}
\begin{tabular}{lcccccccc}
\toprule
\makecell[l]{Dataset \\ Name} & 
\texttt{3drd} & 
\texttt{pol} & 
\texttt{prot} & 
\texttt{bike} & 
\texttt{buzz} \\ 
\midrule
Dim ($D$) & 3 & 26 & 9 & 17 & 77 \\ 
Size ($N$) & 434874 & 15000 & 45730 & 17379 & 583250 \\
Length-\\scale ($l$) & 0.27 & 1.44 & 0.94 & 6.34 & 6.13 \\
Signal\\variance ($\sigma_f^2$) & 0.63 & 0.40 & 0.88 & 7.21 & 1.60 \\
Noise\\variance ($\sigma_n^2$) & 0.31 & 0.04 & 0.33 & 0.07 & 0.09 \\
\bottomrule
\end{tabular}\label{tab:datasets_main_text}
\end{table}
}

The average initial distances of the warm-start initializations are compared with those of the cold-start. Additionally, we compare the number of iterations to convergence for warm and cold-start for the three linear solvers CG, SGD and AP. Convergence is reached when the relative residual norm (residual norm $\lVert\vb{H} \vb*{v} - \vb*{b}\rVert$ divided by $\lVert\bm{b}\rVert$) of the solution is within a tolerance of $\tau=0.01$, which \citet{maddox2021iterativegaussianprocessesreliably} recommends as reliably accurate for predictions with iterative GPs.

\paragraph{Method} We randomly sample 1000 data points $(\vb{X}_1, \vb*{y}_1)$ from a dataset. We optimize the hyperparameters $\vb*{\theta}$ of a Gaussian process with a Mat\'ern-$\frac{3}{2}$ kernel to fit the data by maximizing the marginal log-likelihood (MLL) until the MLL gradient norm is below 0.001. We then compute the smaller kernel matrix $\vb{H}_{11}$ from $\vb{X}_1$ and $\bm{\theta}$ and solve the smaller posterior mean system $\vb{H}_{11}\vb*{u}_1 = \vb*{y}_1$ exactly.
100 additional data points $(\vb{X}_2, \vb*{y}_2)$ are then randomly sampled from the dataset to obtain $\vb{X} = [\vb{X}_1, \vb{X}_2]^\top$ and $\vb*{y} = [\vb*{y}_1, \vb*{y}_2]^\top$.
We then compute the complete kernel matrix $\vb{H}$ from $\vb{X}$ and $\bm{\theta}$ and solve the extended posterior mean system

\begin{equation}
    \vb{H}\vb*{v}=\left[\begin{matrix}\vb{H}_{11}&\vb{H}_{12}\\\vb{H}_{12}^\top&\vb{H}_{22}\\\end{matrix}\right]\begin{bmatrix}
\bm{v}_1 \\
\bm{v}_2
\end{bmatrix}
=
\begin{bmatrix}
\vb*{y}_1 \\
\vb*{y}_2
\end{bmatrix}
= \bm{y}
\label{eq:extended_system_main_text}
\end{equation}

using each of the three iterative solvers. By repeating this experiment with each of the three initializations and the cold-start case as a baseline, we can compare the iterations required for convergence by each method.

\newcommand{\cmrfont}{\fontfamily{cmr}\selectfont}

\begin{figure}
\begin{tikzpicture}[
            every node/.style={font=\fontfamily{cmr}\selectfont},
            every label/.style={font=\fontfamily{cmr}\selectfont}]
    \begin{axis}[
        ybar, 
        title = {\fontfamily{cmr}\selectfont Initial Distance to Solution},
        bar width=0.2cm, 
        symbolic x coords={\texttt{3drd}, \texttt{pol}, \texttt{pro}, \texttt{bike}, \texttt{buzz}}, 
        xtick=data, 
        xticklabel style={rotate=45, anchor=east, text height=1.5ex, text depth=0.25ex},
        xlabel style={yshift=-8pt},
        axis x line=bottom,   
        axis y line*=left,               
        axis line style={-},
        ymajorgrids=true,      
        grid style={solid,gray!30}, 
        ylabel={\fontfamily{cmr}\selectfont Relative Distance}, 
        yticklabel={\fontfamily{cmr}\selectfont\pgfmathprintnumber{\tick}\%},
        ylabel style={yshift=3pt},
        ymin=0, 
        ymax=100, 
        height=4.5cm,
        width=8cm,
        enlarge y limits = -10, 
        enlarge x limits=0.15, 
        legend style={at={(0.5,0.6)}, anchor=south, legend columns=-1, draw=none}, 
    ]
    \addplot+[
        blue, fill=blue!50,
        error bars/.cd,
        y dir=both, 
        y explicit 
    ] 
    coordinates {
        (\texttt{3drd}, 100) +- (0, 0)
        (\texttt{pol}, 100) +- (0, 0)
        (\texttt{pro}, 100) +- (0, 0)
        (\texttt{bike}, 100) +- (0, 0)
        (\texttt{buzz}, 100) +- (0, 0)
    };
    
    \addplot+[
        red, fill=red!50,
        error bars/.cd,
        y dir=both,
        y explicit
    ] 
    coordinates {
        (\texttt{3drd}, 29.7) +- (2.6, 2.6)
        (\texttt{pol}, 28.6) +- (1.4, 1.4)
        (\texttt{pro}, 30.7) +- (1.6, 1.6)
        (\texttt{bike}, 28.2) +- (3.6, 3.6)
        (\texttt{buzz}, 30.2) +- (2.8, 2.8)
    };
    \addplot+[
        custom_green, fill=custom_green!50,
        error bars/.cd,
        y dir=both,
        y explicit
    ] 
    coordinates {
        (\texttt{3drd}, 20.8) +- (1.6, 1.6)
        (\texttt{pol}, 23.5) +- (1.1, 1.1)
        (\texttt{pro}, 21.8) +- (3.2, 3.2)
        (\texttt{bike}, 27.8) +- (3.3, 3.3)
        (\texttt{buzz}, 28.5) +- (2.5, 2.5)
    };\addplot+[
        custom_yellow, fill=custom_yellow!50,
        error bars/.cd,
        y dir=both,
        y explicit
    ] 
    coordinates {
        (\texttt{3drd}, 19.0) +- (1.7, 1.7)
        (\texttt{pol}, 21.9) +- (0.9, 0.9)
        (\texttt{pro}, 16.3) +- (1.3, 1.3)
        (\texttt{bike}, 24.2) +- (3.0, 3.0)
        (\texttt{buzz}, 22.9) +- (2.2, 2.2)
    };
    \end{axis}
\end{tikzpicture}
\begin{center}
    \vspace{-0.3cm}
\hspace{0.5cm}\fontfamily{cmr}\selectfont Dataset
\end{center}
\begin{tikzpicture}
\matrix[column sep=0.2cm, row sep=0cm] {
  \draw[line width=2pt, blue] (0,0) -- (0.4,0) node[right, black, text height=1.5ex, text depth=.25ex]{\fontfamily{cmr}\selectfont Cold-start}; &
  \draw[line width=2pt, red] (0,0) -- (0.4,0) node[right, black, text height=1.5ex, text depth=.25ex]{\fontfamily{cmr}\selectfont Na\"ive warm-start}; \\
  \draw[line width=2pt, custom_green] (0,0) -- (0.4,0) node[right, black, text height=1.5ex, text depth=.25ex]{\fontfamily{cmr}\selectfont Residual line search}; &
  \draw[line width=2pt, custom_yellow] (0,0) -- (0.4,0) node[right, black, text height=1.5ex, text depth=.25ex]{\fontfamily{cmr}\selectfont Marginal system solve}; \\
};
\end{tikzpicture}
\caption{Results for the posterior mean system -- the initial distance of the warm-start initializations are shown as a percentage of the cold-start distance, with mean and standard deviation across trials. The warm-start methods consistently reduce the initial distance by 70-80\% across all datasets, for the setting of 1000 initial data points + 100 new data points.}\label{fig:initial_distance_mean}
\end{figure}

\input{Section_3/plot_relative_iterations}

The experiment is repeated over 10 trials for each dataset to observe the mean and standard deviation of the iterations required for convergence.
Furthermore, the initial RKHS distances

 \begin{equation}
     \left\|\left(\vb*{v} - \vb*{v}^*\right)\right\|_{\vb{H}}=\sqrt{\left(\vb*{v} - \vb*{v}^*\right)^\top\vb{H}\left(\vb*{v} - \vb*{v}^*\right)}
 \end{equation}
 
to the solution are compared.

The same experiment is repeated for an additional 10 trials for each dataset with the posterior sample systems:
\begin{equation}
\vb{H}\vb*{v}
= f\!\left(
\begin{bmatrix}
\vb{X}_1 \\
\vb{X}_2
\end{bmatrix}
\right) + \begin{bmatrix}
\vb*{\epsilon}_1 \\
\vb*{\epsilon}_2
\end{bmatrix}
= f(\vb{X})+\vb*{\epsilon}
\end{equation}

where $f(\cdot)$ is a random sample from the GP prior obtained using 2000 random Fourier features -- an technique introduced in \citet{random_fourier} for approximating kernel methods, with the number of features chosen following the analysis in \citet{sutherland2015error} -- and $\vb*{\epsilon} \sim \mathcal{N}(\vb{0}, \sigma_n^2 \mathbf{I})$ is a vector of random noise samples.

Detailed description of datasets and implementation of linear solvers are provided in \Cref{appendix:implementation}.

%% file: Section_3/plot_relative_iterations.tex
\begin{figure*}[!t]
\begin{center}
    \vspace{-0.3cm}
\hspace{0.5cm}\fontfamily{cmr}\selectfont Linear Solver Iterations for Convergence
\end{center}
\begin{minipage}[t]{0.365\textwidth}
    \begin{tikzpicture}[
            every node/.style={font=\fontfamily{cmr}\selectfont},
            every label/.style={font=\fontfamily{cmr}\selectfont}]
    \begin{axis}[
        ybar, 
        title = {\fontfamily{cmr}\selectfont SGD},
        bar width=0.15cm, 
        symbolic x coords={\texttt{3drd}, \texttt{pol}, \texttt{pro}, \texttt{bike}, \texttt{buzz}}, 
        xtick=data, 
        xticklabel style={rotate=45, anchor=east, text height=1.5ex, text depth=0.25ex},
        xlabel style={yshift=-8pt},
        axis x line=bottom,   
        axis y line*=left,               
        axis line style={-},
        ymajorgrids=true,      
        grid style={solid,gray!30}, 
        ylabel={\fontfamily{cmr}\selectfont Relative Iterations (\%)}, 
        yticklabel={\fontfamily{cmr}\selectfont\pgfmathprintnumber{\tick}},
        ylabel style={yshift=-8pt},
        ymin=0, 
        ymax=100, 
        height=4.5cm,
        width=6.6cm,
        enlarge y limits = -10, 
        enlarge x limits=0.15, 
        legend style={at={(0.5,0.6)}, anchor=south, legend columns=-1, draw=none}, 
    ]
    \addplot+[
        blue, fill=blue!50,
        error bars/.cd,
        y dir=both, 
        y explicit 
    ] 
    coordinates {
        (\texttt{3drd}, 100) +- (0, 0)
        (\texttt{pol}, 100) +- (0, 0)
        (\texttt{pro}, 100) +- (0, 0)
        (\texttt{bike}, 100) +- (0, 0)
        (\texttt{buzz}, 100) +- (0, 0)
    };
    
    \addplot+[
        red, fill=red!50,
        error bars/.cd,
        y dir=both,
        y explicit
    ] 
    coordinates {
        (\texttt{3drd}, 71.9) +- (3.7, 3.7)
        (\texttt{pol}, 59.2) +- (5.3, 5.3)
        (\texttt{pro}, 73.2) +- (12.4, 12.4)
        (\texttt{bike}, 65.3) +- (4.3, 4.3)
        (\texttt{buzz}, 68.0) +- (2.5, 2.5)
    };
    \addplot+[
        custom_green, fill=custom_green!50,
        error bars/.cd,
        y dir=both,
        y explicit
    ] 
    coordinates {
        (\texttt{3drd}, 64.4) +- (2.6, 2.6)
        (\texttt{pol}, 56.7) +- (4.5, 4.5)
        (\texttt{pro}, 67.6) +- (11.2, 11.2)
        (\texttt{bike}, 64.4) +- (3.9, 3.9)
        (\texttt{buzz}, 65.7) +- (2.3, 2.3)
    };\addplot+[
        custom_yellow, fill=custom_yellow!50,
        error bars/.cd,
        y dir=both,
        y explicit
    ] 
    coordinates {
        (\texttt{3drd}, 63.1) +- (3.0, 3.0)
        (\texttt{pol}, 55.2) +- (5.1, 5.1)
        (\texttt{pro}, 62.8) +- (9.4, 9.4)
        (\texttt{bike}, 59.3) +- (4.0, 4.0)
        (\texttt{buzz}, 60.3) +- (2.4, 2.4)
    };
    \end{axis}

\end{tikzpicture}
\end{minipage}
\hfill
\begin{minipage}[t]{0.31\textwidth}
    \begin{tikzpicture}[
            every node/.style={font=\fontfamily{cmr}\selectfont},
            every label/.style={font=\fontfamily{cmr}\selectfont}]
    \begin{axis}[
        ybar, 
        title = {\fontfamily{cmr}\selectfont AP},
        bar width=0.15cm, 
        symbolic x coords={\texttt{3drd}, \texttt{pol}, \texttt{pro}, \texttt{bike}, \texttt{buzz}}, 
        xtick=data, 
        xticklabel style={rotate=45, anchor=east, text height=1.5ex, text depth=0.25ex},
        xlabel style={yshift=-8pt},
        axis x line=bottom,   
        axis y line*=left,               
        axis line style={-},
        y axis line style={draw=none},
        ymajorgrids=true,      
        grid style={solid,gray!30}, 
        yticklabels={},
        ylabel style={yshift=3pt},
        ymin=0, 
        ymax=100, 
        height=4.5cm,
        width=6.6cm,
        enlarge y limits = -10, 
        enlarge x limits=0.15, 
        legend style={at={(0.5,0.6)}, anchor=south, legend columns=-1, draw=none}, 
    ]
    \addplot+[
        blue, fill=blue!50,
        error bars/.cd,
        y dir=both, 
        y explicit 
    ] 
    coordinates {
        (\texttt{3drd}, 100) +- (0, 0)
        (\texttt{pol}, 100) +- (0, 0)
        (\texttt{pro}, 100) +- (0, 0)
        (\texttt{bike}, 100) +- (0, 0)
        (\texttt{buzz}, 100) +- (0, 0)
    };
    
    \addplot+[
        red, fill=red!50,
        error bars/.cd,
        y dir=both,
        y explicit
    ] 
    coordinates {
        (\texttt{3drd}, 41.2) +- (4.1, 4.1)
        (\texttt{pol}, 29.0) +- (3.3, 3.3)
        (\texttt{pro}, 30.5) +- (9.2, 9.2)
        (\texttt{bike}, 24.6) +- (14.9, 14.9)
        (\texttt{buzz}, 8.0) +- (4.3, 4.3)
    };
    \addplot+[
        custom_green, fill=custom_green!50,
        error bars/.cd,
        y dir=both,
        y explicit
    ] 
    coordinates {
        (\texttt{3drd}, 40.4) +- (4.6, 4.6)
        (\texttt{pol}, 28.8) +- (2.8, 2.8)
        (\texttt{pro}, 29.7) +- (8.5, 8.5)
        (\texttt{bike}, 10.5) +- (6.7, 6.7)
        (\texttt{buzz}, 7.7) +- (4.1, 4.1)
    };\addplot+[
        custom_yellow, fill=custom_yellow!50,
        error bars/.cd,
        y dir=both,
        y explicit
    ] 
    coordinates {
        (\texttt{3drd}, 41.2) +- (4.4, 4.4)
        (\texttt{pol}, 29.3) +- (2.5, 2.5)
        (\texttt{pro}, 29.9) +- (9.0, 9.0)
        (\texttt{bike}, 10.7) +- (6.7, 6.7)
        (\texttt{buzz}, 8.0) +- (4.3, 4.3)
    };
    \end{axis}

\end{tikzpicture}
\end{minipage}
\hfill
\begin{minipage}[t]{0.31\textwidth}
\begin{tikzpicture}[
            every node/.style={font=\fontfamily{cmr}\selectfont},
            every label/.style={font=\fontfamily{cmr}\selectfont}]
\begin{axis}[
        ybar, 
        title = {\fontfamily{cmr}\selectfont CG},
        bar width=0.15cm, 
        symbolic x coords={\texttt{3drd}, \texttt{pol}, \texttt{pro}, \texttt{bike}, \texttt{buzz}}, 
        xtick=data, 
        xticklabel style={rotate=45, anchor=east, text height=1.5ex, text depth=0.25ex},
        xlabel style={yshift=-8pt},
        axis x line=bottom,   
        axis y line*=left,               
        axis line style={-},
        y axis line style={draw=none},
        ymajorgrids=true,      
        grid style={solid,gray!30}, 
        yticklabels={},
        ylabel style={yshift=3pt},
        ymin=0, 
        ymax=100, 
        height=4.5cm,
        width=6.6cm,
        enlarge y limits = -10, 
        enlarge x limits=0.15, 
        legend style={at={(0.5,0.6)}, anchor=south, legend columns=-1, draw=none}, 
    ]
    \addplot+[
        blue, fill=blue!50,
        error bars/.cd,
        y dir=both, 
        y explicit 
    ] 
    coordinates {
        (\texttt{3drd}, 100) +- (0, 0)
        (\texttt{pol}, 100) +- (0, 0)
        (\texttt{pro}, 100) +- (0, 0)
        (\texttt{bike}, 100) +- (0, 0)
        (\texttt{buzz}, 100) +- (0, 0)
    };
    
    \addplot+[
        red, fill=red!50,
        error bars/.cd,
        y dir=both,
        y explicit
    ] 
    coordinates {
        (\texttt{3drd}, 70.2) +- (4.1, 4.1)
        (\texttt{pol}, 71.7) +- (3.2, 3.2)
        (\texttt{pro}, 75.2) +- (8.1, 8.1)
        (\texttt{bike}, 64.0) +- (3.4, 3.4)
        (\texttt{buzz}, 72.2) +- (11.7, 11.7)
    };
    \addplot+[
        custom_green, fill=custom_green!50,
        error bars/.cd,
        y dir=both,
        y explicit
    ] 
    coordinates {
        (\texttt{3drd}, 69.3) +- (4.1, 4.1)
        (\texttt{pol}, 67.5) +- (2.7, 2.7)
        (\texttt{pro}, 76.2) +- (8.6, 8.6)
        (\texttt{bike}, 82.6) +- (7.4, 7.4)
        (\texttt{buzz}, 77.9) +- (7.2, 7.2)
    };\addplot+[
        custom_yellow, fill=custom_yellow!50,
        error bars/.cd,
        y dir=both,
        y explicit
    ] 
    coordinates {
        (\texttt{3drd}, 67.5) +- (3.1, 3.1)
        (\texttt{pol}, 66.9) +- (1.8, 1.8)
        (\texttt{pro}, 73.4) +- (8.1, 8.1)
        (\texttt{bike}, 64.1) +- (3.8, 3.8)
        (\texttt{buzz}, 67.2) +- (7.0, 7.0)
    };
    \end{axis}
\end{tikzpicture}
\end{minipage}
\begin{center}
    \vspace{-0.3cm}
\hspace{0.5cm}\fontfamily{cmr}\selectfont Dataset
\end{center}
\input{legend}
\caption{Results for the posterior mean system -- the number of solver iterations required for convergence for the warm-start methods are shown as a percentage of the iterations required for cold-starting, with mean and standard deviation across trials. The smaller initial distance generally results in fewer iterations for all solvers.}\label{fig:solver_iterations_mean}
\end{figure*}

%% file: legend.tex
\begin{center}
 \begin{tikzpicture}
\matrix[column sep=0.5cm, row sep=0.1cm] {
  \draw[line width=2pt, blue] (0,0) -- (0.4,0) node[right, black, text height=1.5ex, text depth=.25ex]{\fontfamily{cmr}\selectfont Cold-start}; &
  \draw[line width=2pt, red] (0,0) -- (0.4,0) node[right, black, text height=1.5ex, text depth=.25ex]{\fontfamily{cmr}\selectfont Na\"ive warm-start}; &
  \draw[line width=2pt, custom_green] (0,0) -- (0.4,0) node[right, black, text height=1.5ex, text depth=.25ex]{\fontfamily{cmr}\selectfont Residual line search}; &
  \draw[line width=2pt, custom_yellow] (0,0) -- (0.4,0) node[right, black, text height=1.5ex, text depth=.25ex]{\fontfamily{cmr}\selectfont Marginal system solve}; \\
};
\end{tikzpicture}
\end{center}

%% file: Section_3/bayes_opt_results.tex
\begin{figure*}[!t]
    \centering
    \includegraphics[width=1\linewidth]{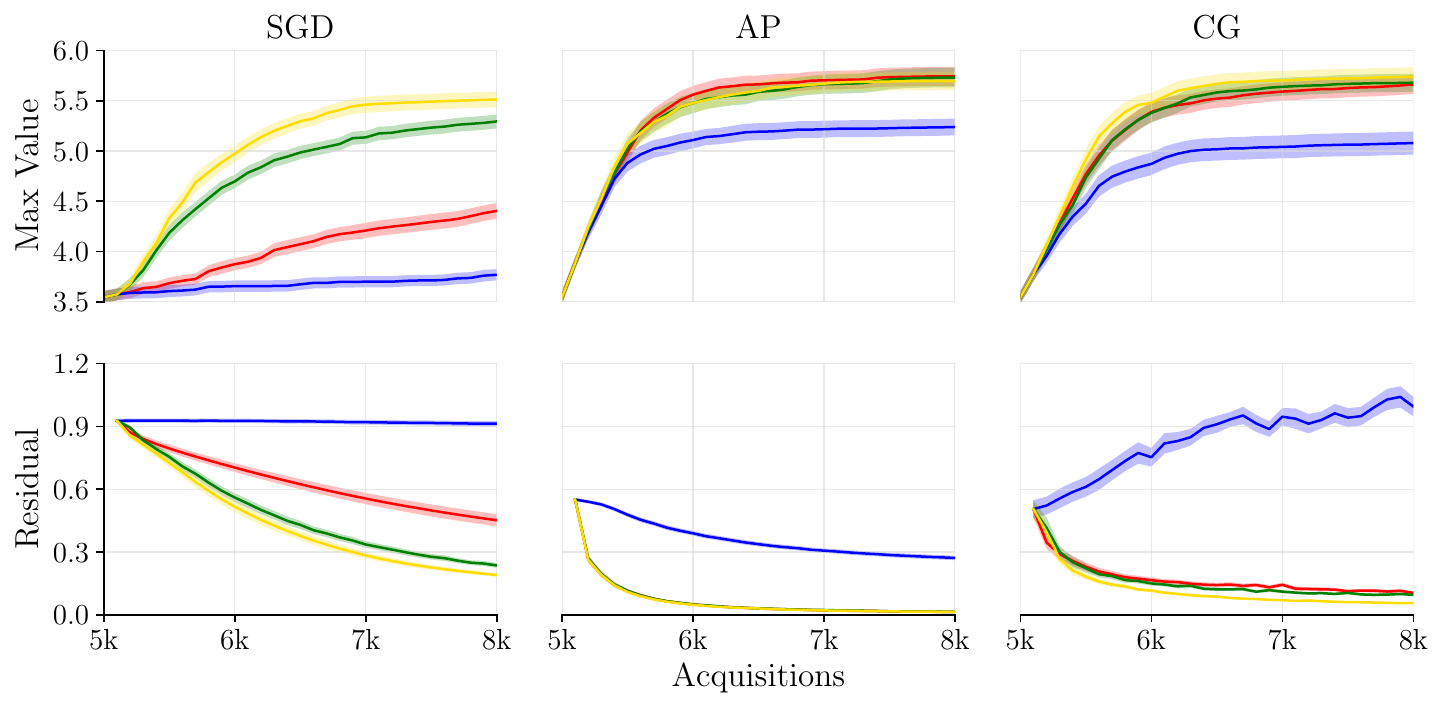}
    
\begin{tikzpicture}
\end{tikzpicture}
\vspace{-20pt}
\input{legend}
    \caption{Maximum objective function values and final residuals of the posterior mean solve, as a function of data point acquisitions. Initialising the weights with more accurate estimates achieves improved Bayesian optimisation performance, reflected in the smaller final residuals. The evolution of the residuals shows that linear solver progress accumulates over multiple solves with warm-starting, rather than resetting after every solve. For SGD, the evolution as points are acquired clearly show that more accurate initialisations result in improved performance, which agrees with our theoretical analysis; for CG and AP, the na\"ive warm-start method is sufficiently accurate under the given compute budget that the performance gains achieved with the more accurate methods are more marginal.}
    \label{fig:small budget}
\end{figure*}

%% file: Section_3/bayes_opt_experiment.tex
In Bayesian optimization (BO), expensive black-box functions are optimized by employing Gaussian processes as surrogate models. Fundamentally, BO relies on accurate posterior predictions and uncertainty estimates, making them a natural benchmark for sequential decision-making. To demonstrate the effectiveness of warm-starting GP posteriors with additional data, a parallel Thompson sampling task similar to the experiment done in \citet{sampling_from_gp_posteriors} is performed. Introduced by \citet{parallel_thompson}, parallel Thompson sampling is an acquisition function for BO which selects new evaluation locations on the objective function by drawing a large number of posterior samples in parallel.


\citet{sampling_from_gp_posteriors} shows that early stopping (limiting the compute budget) of the linear solves can result in much faster GP posterior updates without a substantial loss in Bayesian optimization performance. This reduces the cost of GPs for sequential inference. In this experiment, we use this as a baseline against which to compare our method, demonstrating that our methods for finding more accurate initial estimates of the weights result in improved Bayesian optimization performance under the same limited compute budget.

We evaluate our initialization methods for improving Bayesian optimization performance for three linear solvers CG, SGD and AP, setting the maximum number of solver iterations -- 5, 120 and 30 respectively -- such that each solver runs for the same amount of time.

\paragraph{Method} We follow the experimental protocol from \citet{sampling_from_gp_posteriors}. The objective function to be maximized is a sample drawn from a GP prior $g \sim \mathcal{GP}(0, k)$ with input space $[0,1]^8$. The ground-truth hyperparameters are then used in parallel Thompson sampling to optimize the objective function, which prevents model misspecification confounding in the results by isolating the effect of solver initialization methods from hyperparameter tuning \citep{sampling_from_gp_posteriors}.

We start with 5000 initial points, which are spaced uniformly in the input domain. The objective function is computed at these initial points (with Gaussian observation noise): $y=g(\bm{x})+\epsilon$. These points $(\bm{x},y)$ are added to the training data.

Then for each batch of acquisitions, 100 samples from the GP posterior distribution are drawn using pathwise conditioning. The input locations which maximize each posterior sample are found by performing gradient ascent separately on each posterior sample starting from a few initial locations (full details in \Cref{appendix:parallelThompson}). These inputs are chosen for evaluation on the objective function; the 100 new objective function values are then added to the training data.

Updating the posterior samples after adding each batch of points to the training dataset requires solving the extended linear systems. To compare the warm and cold-start initializations, we applied each to a parallel Thompson experiment using one of 5 kernel lengthscales \{0.1, 0.2, 0.3, 0.4, 0.5\} with the Mat\'ern-$\frac{3}{2}$ kernel, ensuring robustness of results across different kernels. We also used a signal variance of 1.0 and a noise-scale of 0.001. The Mat\'ern kernel with the smoothness parameter $\nu=\frac{3}{2}$ is used for its ability to model functions with rough jump-like behavior \citep{matern2013spatial}.

For each of the 5 lengthscales, we ran the experiment with 10 random seeds, giving 50 trials for each of the 4 initializations with the mean and standard error over those 50 trials shown in \Cref{fig:small budget}. This is repeated for each of the 3 linear solvers, resulting in a total of 600 runs and a cumulative runtime of 471 GPU hours on Nvidia GeForce RTX 2080 Ti GPUs.

%% file: conclusion.tex
This paper introduces the use of warm-start methods for iterative linear solvers to improve and speed-up the computation of Gaussian process posteriors in sequential settings. For the linear solves that arise when additional data points are added to update the posterior model, we have demonstrated both theoretically and empirically that our three proposed methods reduce the initial distance to the final solution. 
Through regression experiments on real-world datasets, we have shown that when solving until the residuals converge, our methods for warm-starting are effective at speeding-up solves by an average of 1.4$\times$ for CG, 1.6$\times$ for SGD and 4.2$\times$ for AP. \marker{black} In a parallel Thompson sampling experiment, we found that under a limited compute setting, our methods improve convergence, which results in more accurate posterior predictions and better Bayesian optimization performance; our methods improved the final maximum found on the objective function by up to 13\% for CG, 46\% for SGD and 10\% for AP. The results demonstrate that warm-starting is a practical and effective strategy for applications of GPs in sequential inference. Overall, this paper contributes to improving the scalability of Gaussian processes by reducing the cost of updating posteriors, a common limiting factor constraining the performance of GPs in large-scale probabilistic modeling tasks.

%% file: Appendices/derivation.tex
Let $\vb*{u}_1$ denote the known solution to a linear system with $n_1$ data points: 

\begin{equation}
    \vb{H}_{11}\vb*{u}_1 = \vb*{b}_1.
\end{equation}


Introducing $n_2$ additional data points gives an extended linear system which contains the matrix $\vb{H}_{11}$ and the right-hand side vector $\bm{b}_1$ from the original system:

\begin{equation}
    \vb{H}\vb*{v}=\left[\begin{matrix}\vb{H}_{11}&\vb{H}_{12}\\\vb{H}_{12}^\top&\vb{H}_{22}\\\end{matrix}\right]\begin{bmatrix}
\bm{v}_1 \\
\bm{v}_2
\end{bmatrix}
=
\begin{bmatrix}
\vb*{b}_1 \\
\vb*{b}_2
\end{bmatrix}
= \bm{b}.
\label{eq:joint_system}
\end{equation}

Iterative solvers find the weights $\bm{v}_1$ and $\bm{v}_2$ which solve the extended linear system by minimizing the quadratic objective
\begin{equation}
\begin{gathered}
J\left(\bm{v}\right)=J(\bm{v}_1,\bm{v}_2)\; =\;\tfrac12\begin{bmatrix}\bm{v}_1\\\bm{v}_2\end{bmatrix}^{\!\top}
\begin{bmatrix}\vb{H}_{11} & \vb{H}_{12}\\ \vb{H}_{12}^\top & \vb{H}_{22}\end{bmatrix}
\begin{bmatrix}\bm{v}_1\\\bm{v}_2\end{bmatrix}
- \begin{bmatrix}\bm{b}_1\\\bm{b}_2\end{bmatrix}^{\!\top}\!
\begin{bmatrix}\bm{v}_1\\\bm{v}_2\end{bmatrix} \\
    =\tfrac12 \bm{v}_1^\top \vb{H}_{11} \bm{v}_1 + \bm{v}_1^\top \vb{H}_{12} \bm{v}_2 + \tfrac12 \bm{v}_2^\top \vb{H}_{22}\bm{v}_2 - \bm{b}_1^\top \bm{v}_1 - \bm{b}_2^\top \bm{v}_2.
\end{gathered}
\label{eq:extended_objective}
\end{equation}

\subsection{How the quadratic objective relates to the initial RKHS-norm distance and the initial residual}

Let $\bm{v}^{(i)}$ denote the values of the weights after the i-th solver iteration, and accordingly $\bm{v}^{(0)}$ the values used to initialize an iterative linear solver. 
The RKHS distance of the initialization from the exact solution is

\begin{equation}
    d^2=\left\|\bm{v}^{(0)}-\bm{v}^*\right\|^2_{\vb{H}}=\left(\bm{v}^{(0)}-\bm{v}^*\right)^\top \vb{H} \left(\bm{v}^{(0)}-\bm{v}^*\right),
\end{equation}

where $\bm{v}^*$ is the exact solution of the extended linear system.

Similarly, the residual of the linear system after the i-th solver iteration, the norm of which is used as a stopping criterion for the linear solvers, is defined as

\begin{equation}
    \bm{r}^{(i)} = \vb{H}\bm{v}^{(i)}-\bm{b}=\vb{H}\left(\bm{v}^{(i)}-\bm{v}^*\right).
\end{equation}

 The difference between the initial value and the minimum of the quadratic objective can be expressed both in terms of the initial squared RKHS ($\vb{H}$-norm) distance and squared $\vb{H}^{-1}$-norm of the initial residual:
 
\begin{equation}
\begin{gathered}
J\left(\bm{v}^{(0)}\right)-J(\bm{v}^*)=\tfrac12 \bm{v}^{(0)\top} \vb{H} \bm{v}^{(0)} - \bm{b}^\top \bm{v}^{(0)} - \left(\tfrac12 \bm{v}^{*\top} \vb{H} \bm{v}^*- \bm{b}^\top \bm{v}^*\right)
\\
=\tfrac12 \bm{v}^{(0)\top} \vb{H} \bm{v}^{(0)} - \bm{v}^{*\top}\vb{H}^\top \bm{v}^{(0)} - \tfrac12 \bm{v}^{*\top} \vb{H} \bm{v}^* + \bm{v}^{*\top}\vb{H} \bm{v}^*
=\tfrac12\left(\bm{v}^{(0)}-\bm{v}^*\right)^\top \vb{H} \left(\bm{v}^{(0)}-\bm{v}^*\right)\\=\tfrac12d^2
\;=\;\tfrac12\,\bm{r}^{(0)\top} \vb{H}^{-1} \bm{r}^{(0)},
\label{eq:equivalency}
\end{gathered}
\end{equation}

where we use the fact that $\bm{b}=\vb{H}\bm{v}^*$ and $\vb{H}=\vb{H}^\top$.
Hence an improvement (decrease) of the quadratic objective is equivalent to a reduction of the distance of the weights to the solution (RKHS norm) and a reduction of the residual. A change in the squared RKHS distance to the solution varies linearly with a change in the value of the quadratic objective:
\begin{equation}
    \Delta d^2=2\Delta J.
\end{equation}

\subsection[All-zero (cold-start) initialization]{All-zero (cold-start) initialization: $(\bm{v}_1,\bm{v}_2)=(\bm{0},\bm{0})$}

The objective (\ref{eq:extended_objective}) evaluated at the cold-start initialization is

\begin{equation}
  J(\bm{0},\bm{0}) = 0.  
\end{equation}

Substituting $\bm{v}^{(0)}=\bm{0}$ into \eqref{eq:equivalency}, we find that

\begin{equation}
    d^2_{cold} =\bm{v}^{*\top}\vb{H}\bm{v}^{*}=\bm{b}^{\top}\vb{H}^{-1}\bm{b},
\end{equation}
where we use $\bm{v}^{*}=\vb{H}^{-1}\bm{b}$.

\citet{cookbook} gives the inverse of a block matrix in terms $\vb{S}=\vb{H}_{22}-\vb{H}_{12}^\top\vb{H}_{11}^{-1}\vb{H}_{12}$, which is known as the Schur complement for $\vb{H}_{11}$:

\begin{equation}
\left[\begin{matrix}
\vb{H}_{11}&\vb{H}_{12}\\\vb{H}_{12}^\top&\vb{H}_{22}\\\end{matrix}\right]^{-1} = 
\left[\begin{matrix}
\vb{H}_{11}^{-1}+\vb{H}_{11}^{-1}\vb{H}_{12}\vb{S}^{-1}\vb{H}_{12}^\top\vb{H}_{11}^{-1}
&-\vb{H}_{11}^{-1}\vb{H}_{12}\vb{S}^{-1}
\\-\vb{S}^{-1}\vb{H}_{12}^\top\vb{H}_{11}^{-1}
&\vb{S}^{-1}\\\end{matrix}\right].
\end{equation}

Hence for cold-starting, the squared RKHS distance is,
\begin{gather}
d_{cold}^2 =\bm{b}^\top\vb{H}^{-1}\bm{b} \notag 
\\
= \bm{b}_1^\top\vb{H}_{11}^{-1}\bm{b}_1 + \bm{b}_1^\top\vb{H}_{11}^{-1}\vb{H}_{12}\vb{S}^{-1}\vb{H}_{12}^\top\vb{H}_{11}^{-1}\bm{b}_1
-2\bm{b}_1^\top\vb{H}_{11}^{-1}\vb{H}_{12}\vb{S}^{-1}\bm{b}_2
+\bm{b}_2^\top\vb{S}^{-1}\bm{b}_2\label{eq:cold_rkhs}.
\end{gather}

\subsection{Warm-start initializations}

For all three warm-start initializations, \(\bm{v}^{(0)}_1=\bm{u}_1\), so the initial value of the quadratic objective can be simplified to an expression containing an additive constant independent of \(\bm{v}_{2}\):

\begin{equation}
    J(\bm{u}_{1},\bm{v}_{2}) = \text{const} \;+\; \tfrac12 \bm{v}_{2}^\top \vb{H}_{22} \bm{v}_{2} - \bm{v}_{2}^\top \left(\bm{b}_{2} - \vb{H}_{12}^\top \bm{u}_{1}\right),
\end{equation}

where \(\text{const} = \tfrac12 \bm{u}_{1}^\top \vb{H}_{11} \bm{u}_{1} - \bm{b}_{1}^\top \bm{u}_{1}\).

Note: the na\"ive warm-start initialization ($(\bm{v}_1,\bm{v}_2)=(\bm{u}_1,\bm{0})$) leaves a residual of 

\begin{equation}
    \bm{r} =\begin{bmatrix}\bm{b}_1\\\bm{b}_2\end{bmatrix}-\begin{bmatrix}\vb{H}_{11} & \vb{H}_{12}\\ \vb{H}_{12}^\top & \vb{H}_{22}\end{bmatrix}\begin{bmatrix}\bm{u}_1\\\bm{0}\end{bmatrix}
=\bm{b}_2-\vb{H}_{12}^\top \bm{u}_1,
\end{equation}
using the fact that $\bm{b}_1-\vb{H}_{11}\bm{u}_1 = \bm{0}$. We can then further simplify the expression for the objective:

\begin{equation}
    J(\bm{u}_{1},\bm{v}_{2}) = \text{const} \;+\; \tfrac12 \bm{v}_{2}^\top \vb{H}_{22} \bm{v}_{2} - \bm{v}_{2}^\top \bm{r}.
    \label{eq:reduced_objective}
\end{equation}

\subsection[Na\"ive warm-start]{Method 1: Na\"ive warm-start $(\bm{v}_1,\bm{v}_2)=(\bm{u}_1,\bm{0})$}
Substituting $\bm{v}_2=\bm{0}$ into the simplified objective (\ref{eq:reduced_objective}), we get

\begin{equation}
    J(\bm{u}_{1},\bm{0})
= \text{const} = \tfrac12 \bm{u}_{1}^\top \vb{H}_{11} \bm{u}_{1} - \bm{b}_{1}^\top \bm{u}_{1}
= -\tfrac12 \bm{u}_{1}^\top \vb{H}_{11} \bm{u}_{1},
\end{equation}

where we use the fact that $\bm{b}_1 = \vb{H}_{11} \bm{u}_1$ and $\vb{H}_{11}=\vb{H}_{11}^\top$.

Thus the objective improves  (decreases) by
\begin{equation}
J(\bm{0},\bm{0})-J(\bm{u}_{1},\bm{0})
= \tfrac12 \bm{u}_{1}^\top \vb{H}_{11} \bm{u}_{1}
= \tfrac12\, \bm{b}_{1}^\top \vb{H}_{11}^{-1} \bm{b}_{1}\geq0.
\end{equation}

Hence, the difference between the initial squared RKHS distance of cold-starting and na\"ive warm-starting is

\begin{equation}
    d_\text{cold}^2-d_\text{na\"ive}^2=\bm{b}_1^\top\vb{H}_{11}^{-1}\bm{b}_1\geq0.
\end{equation}

The reduction of $\bm{b}_1^\top\vb{H}_{11}^{-1}\bm{b}_1$ is also the first term present in \eqref{eq:cold_rkhs}. The final inequality comes from the fact that $\vb{H}_{11}^{-1}$ is positive definite, with equality only when $\bm{b}_1=\bm{0}$. Hence, warm-starting strictly reduces the initial distance to the solution. In general, this reduction will also be a greater proportion of the cold-start distance if $n_1$ is larger relative to $n_2$, since the $\bm{b}_1^\top\vb{H}_{11}^{-1}\bm{b}_1$ term is more dominant compared to the other terms in \eqref{eq:cold_rkhs}.

\subsection[Method 2]{Method 2: Residual line search $(\bm{v}_1,\bm{v}_2)=\left(\bm{u}_1, \frac{\bm{r}^\top\bm{r}}{\bm{r}^\top\vb{H}_{22}\bm{r}}\bm{r}\right)$}

Instead of the na\"ive method of initializing $\bm{v}_2$ at the origin, we propose to initialize it as an optimally scaled vector in the direction of the residual, i.e., \(\bm{v}_2=\alpha \bm{r}\).

Substituting this one-parameter form into the objective \eqref{eq:reduced_objective} yields

\begin{equation}
  J(\bm{u}_1,\alpha \bm{r})
= \text{const} + \tfrac12 \alpha^2 \bm{r}^\top \vb{H}_{22} \bm{r} - \alpha\bm{r}^\top \bm{r}.
\end{equation}

The value of $\alpha$ which minimizes the objective is found by differentiating w.r.t. $\alpha$:

\begin{equation}
    \frac{\partial J(\bm{u}_1,\alpha \bm{r})}{\partial\alpha}=\alpha \bm{r}^\top \vb{H}_{22} \bm{r} - \bm{r}^\top \bm{r}
\end{equation}
\begin{equation}
    \Rightarrow \alpha^\star \;=\; \frac{\bm{r}^\top \bm{r}}{\,\bm{r}^\top \vb{H}_{22} \bm{r}\,},
\end{equation}

which resembles an inverse Rayleigh quotient (see \citet{golub2013matrix}).

The objective with this method becomes
\begin{equation}
    J(\bm{u}_1,\alpha^\star \bm{r}) = J(\bm{u}_1,\bm{0}) - \frac{1}{2}\,\frac{(\bm{r}^\top \bm{r})^2}{\,\bm{r}^\top \vb{H}_{22} \bm{r}\,}.
\end{equation}

Hence the improvement over the na\"ive warm-start method is

\begin{equation}
J(\bm{u}_1,\bm{0}) - J(\bm{u}_1,\alpha^\star \bm{r})
= \frac{1}{2}\,\frac{(\bm{r}^\top \bm{r})^2}{\,\bm{r}^\top \vb{H}_{22} \bm{r}\,}>0
\end{equation}

in terms of the objective and

\begin{equation}
d_\text{na\"ive}^2-d_\text{line-search}^2
= \frac{(\bm{r}^\top \bm{r})^2}{\,\bm{r}^\top \vb{H}_{22} \bm{r}\,}>0
\end{equation}

in terms of the RKHS distance. The inequality is true since $\vb{H}_{22}$ is positive definite.

\subsection[Method 3]{Method 3: Marginal solve \((\bm{v}_1,\bm{v}_2)=(\bm{u}_1, \vb{H}_{22}^{-1}\bm{r})\)}

Again using the simplified objective \eqref{eq:reduced_objective}, we instead differentiate w.r.t. $\bm{v}_2$, finding the weights $\bm{v}_2$ which minimize the objective given that $\bm{v}_1$ is fixed:

\begin{equation}
    \frac{\partial J(\bm{u}_1,\bm{v}_2)}{\partial\bm{v}_2}=\vb{H}_{22} \bm{v}_{2} -  \bm{r}
\end{equation}
\begin{equation}
    \Rightarrow \bm{v}_2^\star \;=\; \vb{H}_{22}^{-1}\bm{r}.
\end{equation}

Hence, we can initialize $\bm{v}_2$ by performing a linear solve on the small marginal system above.

Substituting this into the objective we get

\begin{equation}
J(\bm{u}_1,\bm{v}_2^\star)
= J(\bm{u}_1,\bm{0}) - \tfrac12\, \bm{r}^\top \vb{H}_{22}^{-1} \bm{r},
\end{equation}

hence the improvement over the na\"ive warm-start method is found:
\begin{equation}
    J(\bm{u}_1,\bm{0}) - J(\bm{u}_1,\bm{v}_2^\star)
= \tfrac12\, \bm{r}^\top \vb{H}_{22}^{-1} \bm{r}>0.
\end{equation}

The initial distance is therefore reduced:
\begin{equation}
d_\text{na\"ive}^2-d_\text{marginal-solve}^2
=  \tfrac12\, \bm{r}^\top \vb{H}_{22}^{-1} \bm{r}>0,
\end{equation}

where the inequality is again true because $\vb{H}_{22}^{-1}$ is positive definite.

\subsection{Comparison: methods 2 and 3}

Method 3 solves the marginal system $\vb{H}_{22} \bm{v}_{2} = \bm{r}$ exactly to minimise the quadratic objective, whereas the method 2 performs what is equivalent to a steepest descent step on the marginal system (see Chapter 5.3 of \citet{saad2003iterative}). Method 3 therefore achieves the smallest initial distance out of all three proposed initialisations.

We can confirm this using the Cauchy-Schwarz inequality \citep{wu2009various}, which states that:

\begin{equation}
    (\bm{a}_1^\top \bm{a}_2)^2 \le \|\bm{a}_1\|^2 \, \|\bm{a}_2\|^2.
\end{equation}

Substituting $\bm{a}_1 = \vb{H}_{22}^{1/2} \bm{r}, \bm{a}_2 = \vb{H}_{22}^{-1/2} \bm{r}$ into the inequality gives

\begin{equation}
(\bm{r}^\top \bm{r})^2
\le (\bm{r}^\top \vb{H}_{22} \bm{r})\,(\bm{r}^\top \vb{H}_{22}^{-1} \bm{r}),
\end{equation}

which can be rearranged to yield
\begin{equation}
    \frac{(\bm{r}^\top \bm{r})^2}{\bm{r}^\top \vb{H}_{22} \bm{r}} \le \bm{r}^\top \vb{H}_{22}^{-1} \bm{r},
\end{equation}

with equality when \(\bm{r}\) is an eigenvector of \(\vb{H}_{22}\).

Hence, the marginal system solve method is a more accurate initialization than the reduced line search method:
\begin{equation}
d^2_\text{line-search}-d^2_\text{marginal-solve} \geq 0.
\end{equation}

%% file: Appendices/implementation.tex
\paragraph{General} Our code implementation for all experiments uses the \texttt{PyTorch} library \citep{pytorch_neurips}. All experiments were performed with Nvidia GeForce RTX 2080 Ti GPUs in double floating point precision.
\paragraph{Datasets}The regression experiments are conducted with datasets from the popular UC Irvine Machine Learning repository \citep{asuncion2007uci}, which can be accessed under a Creative Commons Attribution 4.0 International (CC BY 4.0) license. The datasets used are listed in \Cref{tab:datasets} with their dimensionality and size.

\begin{table*}[htbp]
\centering
\caption{Dimensionality, size and tuned hyperparameters of datasets used in experiments. The Matérn-$\frac{3}{2}$ kernel hyperparameters are the averages across 10 trials, obtained by optimizing the marginal log-likelihood.}
\vspace{0.5em}
\begin{tabular}{lcccccccc}
\toprule
\makecell[l]{Dataset \\ Name} & 
\texttt{3drd} & 
\texttt{pol} & 
\texttt{prot} & 
\texttt{bike} & 
\texttt{buzz} \\ 
\midrule
Dim ($D$) & 3 & 26 & 9 & 17 & 77 \\ 
Size ($N$) & 434874 & 15000 & 45730 & 17379 & 583250 \\
Length-\\scale ($l$) & 0.27 & 1.44 & 0.94 & 6.34 & 6.13 \\
Signal\\variance ($\sigma_f^2$) & 0.63 & 0.40 & 0.88 & 7.21 & 1.60 \\
Noise\\variance ($\sigma_n^2$) & 0.31 & 0.04 & 0.33 & 0.07 & 0.09 \\
\bottomrule
\end{tabular}\label{tab:datasets}
\end{table*}

\paragraph{Linear solvers} Three iterative linear solvers are investigated in these experiments to solve systems in the form of $\vb{H}\vb*{v} = \vb*{b}$: conjugate gradients, stochastic gradient descent and alternating projections. The algorithms are implemented exactly as described in the pseudo-code in ``Improving Linear System Solvers for Hyperparameter
Optimization in Iterative Gaussian Processes'' \citep{improving_linear_system_solvers}. Specific details for each solver are provided below.

\paragraph{Conjugate gradients}
In line with \citet{exactgaussianprocesses} and \citet{improving_linear_system_solvers}, we apply a pivoted Cholesky preconditioner of rank 100 to improve the condition number of $\vb{H}$ which improves convergence.

\paragraph{Stochastic gradient descent}
Following \citet{lin2023stochastic}, a learning rate, $\eta$ of 0.3 and momentum $\gamma$, of 0.9 are used for the parallel Thompson experiments. A batch size, $b$ of 100 is found to give good convergence. For the regression experiments, an individual learning rate was found for each dataset through a grid search:
\texttt{3drd}: 0.5, \texttt{pol}: 1.5, \texttt{prot}: 0.6, \texttt{bike}: 0.1, \texttt{buzz}: 0.6.

\paragraph{Alternating projections}
We use a block size of 100 for all experiments, which corresponds to the number of new data points added to the extended linear system in all experiments.

%% file: Appendices/regression_results.tex
\begin{figure}[H]
    \centering
    \begin{center}
       \hspace{1cm}\fontfamily{cmr}\selectfont Initial Distance to Solution
    \end{center}
\begin{minipage}[t]{0.4\textwidth}
\begin{tikzpicture}[
            every node/.style={font=\fontfamily{cmr}\selectfont},
            every label/.style={font=\fontfamily{cmr}\selectfont}]
    \begin{axis}[
        ybar, 
        title = {\fontfamily{cmr}\selectfont Posterior mean system: \bf $\vb{H}\bm{v} = \bm{y}$},
        bar width=0.2cm, 
        symbolic x coords={\texttt{3drd}, \texttt{pol}, \texttt{pro}, \texttt{bike}, \texttt{buzz}}, 
        xtick=data, 
        xticklabel style={rotate=45, anchor=east, text height=1.5ex, text depth=0.25ex},
        xlabel style={yshift=-8pt},
        axis x line=bottom,   
        axis y line*=left,               
        axis line style={-},
        ymajorgrids=true,      
        grid style={solid,gray!30}, 
        ylabel={\fontfamily{cmr}\selectfont Relative Distance}, 
        yticklabel={\fontfamily{cmr}\selectfont\pgfmathprintnumber{\tick}\%},
        ylabel style={yshift=3pt},
        ymin=0, 
        ymax=100, 
        height=4cm,
        width=8cm,
        enlarge y limits = -10, 
        enlarge x limits=0.15, 
        legend style={at={(0.5,0.6)}, anchor=south, legend columns=-1, draw=none}, 
    ]
    \addplot+[
        blue, fill=blue!50,
        error bars/.cd,
        y dir=both, 
        y explicit 
    ] 
    coordinates {
        (\texttt{3drd}, 100) +- (0, 0)
        (\texttt{pol}, 100) +- (0, 0)
        (\texttt{pro}, 100) +- (0, 0)
        (\texttt{bike}, 100) +- (0, 0)
        (\texttt{buzz}, 100) +- (0, 0)
    };
    
    \addplot+[
        red, fill=red!50,
        error bars/.cd,
        y dir=both,
        y explicit
    ] 
    coordinates {
        (\texttt{3drd}, 29.7) +- (2.6, 2.6)
        (\texttt{pol}, 28.6) +- (1.4, 1.4)
        (\texttt{pro}, 30.7) +- (1.6, 1.6)
        (\texttt{bike}, 28.2) +- (3.6, 3.6)
        (\texttt{buzz}, 30.2) +- (2.8, 2.8)
    };
    \addplot+[
        custom_green, fill=custom_green!50,
        error bars/.cd,
        y dir=both,
        y explicit
    ] 
    coordinates {
        (\texttt{3drd}, 20.8) +- (1.6, 1.6)
        (\texttt{pol}, 23.5) +- (1.1, 1.1)
        (\texttt{pro}, 21.8) +- (3.2, 3.2)
        (\texttt{bike}, 27.8) +- (3.3, 3.3)
        (\texttt{buzz}, 28.5) +- (2.5, 2.5)
    };\addplot+[
        custom_yellow, fill=custom_yellow!50,
        error bars/.cd,
        y dir=both,
        y explicit
    ] 
    coordinates {
        (\texttt{3drd}, 19.0) +- (1.7, 1.7)
        (\texttt{pol}, 21.9) +- (0.9, 0.9)
        (\texttt{pro}, 16.3) +- (1.3, 1.3)
        (\texttt{bike}, 24.2) +- (3.0, 3.0)
        (\texttt{buzz}, 22.9) +- (2.2, 2.2)
    };
    \end{axis}
\end{tikzpicture}
\end{minipage}
\hspace{1cm}
\begin{minipage}[t]{0.4\textwidth}
\begin{tikzpicture}[
            every node/.style={font=\fontfamily{cmr}\selectfont},
            every label/.style={font=\fontfamily{cmr}\selectfont}]
    \begin{axis}[
        ybar, 
        title ={\fontfamily{cmr}\selectfont Post. sample system: \bf $\vb{H}\bm{v} = f(\vb{X} )+\vb*{\epsilon}$},
        bar width=0.2cm, 
        symbolic x coords={\texttt{3drd}, \texttt{pol}, \texttt{pro}, \texttt{bike}, \texttt{buzz}}, 
        xtick=data, 
        xticklabel style={rotate=45, anchor=east, text height=1.5ex, text depth=0.25ex},
        xlabel style={yshift=-8pt},
        axis x line=bottom,   
        axis y line*=left,               
        axis line style={-},
        y axis line style = {draw=none},
        ymajorgrids=true,      
        grid style={solid,gray!30}, 
        yticklabels = {},
        ylabel style={yshift=3pt},
        ymin=0, 
        ymax=100, 
        height=4cm,
        width=8cm,
        enlarge y limits = -10, 
        enlarge x limits=0.15, 
        legend style={at={(0.5,0.6)}, anchor=south, legend columns=-1, draw=none}, 
    ]
    \addplot+[
        blue, fill=blue!50,
        error bars/.cd,
        y dir=both, 
        y explicit 
    ] 
    coordinates {
        (\texttt{3drd}, 100) +- (0, 0)
        (\texttt{pol}, 100) +- (0, 0)
        (\texttt{pro}, 100) +- (0, 0)
        (\texttt{bike}, 100) +- (0, 0)
        (\texttt{buzz}, 100) +- (0, 0)
    };
    
    \addplot+[
        red, fill=red!50,
        error bars/.cd,
        y dir=both,
        y explicit
    ] 
    coordinates {
        (\texttt{3drd}, 29.7) +- (2.6, 2.6)
        (\texttt{pol}, 28.6) +- (1.4, 1.4)
        (\texttt{pro}, 30.7) +- (1.6, 1.6)
        (\texttt{bike}, 28.2) +- (3.6, 3.6)
        (\texttt{buzz}, 30.2) +- (2.8, 2.8)
    };
    \addplot+[
        custom_green, fill=custom_green!50,
        error bars/.cd,
        y dir=both,
        y explicit
    ] 
    coordinates {
        (\texttt{3drd}, 20.8) +- (1.6, 1.6)
        (\texttt{pol}, 23.5) +- (1.1, 1.1)
        (\texttt{pro}, 21.8) +- (3.2, 3.2)
        (\texttt{bike}, 27.8) +- (3.3, 3.3)
        (\texttt{buzz}, 28.5) +- (2.5, 2.5)
    };\addplot+[
        custom_yellow, fill=custom_yellow!50,
        error bars/.cd,
        y dir=both,
        y explicit
    ] 
    coordinates {
        (\texttt{3drd}, 19.0) +- (1.7, 1.7)
        (\texttt{pol}, 21.9) +- (0.9, 0.9)
        (\texttt{pro}, 16.3) +- (1.3, 1.3)
        (\texttt{bike}, 24.2) +- (3.0, 3.0)
        (\texttt{buzz}, 22.9) +- (2.2, 2.2)
    };
    \end{axis}
\end{tikzpicture}
\end{minipage}
\begin{center}
    \vspace{-0.3cm}
\hspace{0.5cm}\fontfamily{cmr}\selectfont Dataset
\end{center}
\input{legend}
\caption{Initial distances of the warm-start initializations are shown as a percentage of the cold-start distance, with mean and standard deviation across trials.}\label{fig:initial_distance_mean_appendix}
\end{figure}

\vspace{40pt}

\begin{figure}[H]
\begin{center}
    \vspace{-0.3cm}
\hspace{0.5cm}\fontfamily{cmr}\selectfont \quad Linear Solver Iterations for Convergence
\end{center}

\centering
\begin{minipage}[c]{0.02\textwidth}
  \centering
    \hspace{30pt}
  \rotatebox{90}{\fontfamily{cmr}\selectfont Relative Iterations (\%)} 
      \vfill
\end{minipage}%
\begin{minipage}[t]{0.95\textwidth} 
\begin{minipage}[t]{0.345\textwidth}
    \begin{tikzpicture}[
            every node/.style={font=\fontfamily{cmr}\selectfont},
            every label/.style={font=\fontfamily{cmr}\selectfont}]
    \begin{axis}[
        ybar, 
        title = {\parbox[c]{\linewidth}{\centering \fontfamily{cmr}\selectfont SGD\\ \quad}},
        bar width=0.15cm, 
        symbolic x coords={\texttt{3drd}, \texttt{pol}, \texttt{pro}, \texttt{bike}, \texttt{buzz}}, 
        xtick=data, 
        xticklabel style={rotate=45, anchor=east, text height=1.5ex, text depth=0.25ex},
        xlabel style={yshift=-8pt},
        axis x line=bottom,   
        axis y line*=left,               
        axis line style={-},
        ymajorgrids=true,      
        grid style={solid,gray!30}, 
        yticklabel={\fontfamily{cmr}\selectfont\pgfmathprintnumber{\tick}},
        ylabel style={yshift=-8pt},
        xticklabels={},
        ymin=0, 
        ymax=100, 
        height=4cm,
        width=6.6cm,
        enlarge y limits = -10, 
        enlarge x limits=0.15, 
        legend style={at={(0.5,0.6)}, anchor=south, legend columns=-1, draw=none}, 
    ]
    \addplot+[
        blue, fill=blue!50,
        error bars/.cd,
        y dir=both, 
        y explicit 
    ] 
    coordinates {
        (\texttt{3drd}, 100) +- (0, 0)
        (\texttt{pol}, 100) +- (0, 0)
        (\texttt{pro}, 100) +- (0, 0)
        (\texttt{bike}, 100) +- (0, 0)
        (\texttt{buzz}, 100) +- (0, 0)
    };
    
    \addplot+[
        red, fill=red!50,
        error bars/.cd,
        y dir=both,
        y explicit
    ] 
    coordinates {
        (\texttt{3drd}, 71.9) +- (3.7, 3.7)
        (\texttt{pol}, 59.2) +- (5.3, 5.3)
        (\texttt{pro}, 73.2) +- (12.4, 12.4)
        (\texttt{bike}, 65.3) +- (4.3, 4.3)
        (\texttt{buzz}, 68.0) +- (2.5, 2.5)
    };
    \addplot+[
        custom_green, fill=custom_green!50,
        error bars/.cd,
        y dir=both,
        y explicit
    ] 
    coordinates {
        (\texttt{3drd}, 64.4) +- (2.6, 2.6)
        (\texttt{pol}, 56.7) +- (4.5, 4.5)
        (\texttt{pro}, 67.6) +- (11.2, 11.2)
        (\texttt{bike}, 64.4) +- (3.9, 3.9)
        (\texttt{buzz}, 65.7) +- (2.3, 2.3)
    };\addplot+[
        custom_yellow, fill=custom_yellow!50,
        error bars/.cd,
        y dir=both,
        y explicit
    ] 
    coordinates {
        (\texttt{3drd}, 63.1) +- (3.0, 3.0)
        (\texttt{pol}, 55.2) +- (5.1, 5.1)
        (\texttt{pro}, 62.8) +- (9.4, 9.4)
        (\texttt{bike}, 59.3) +- (4.0, 4.0)
        (\texttt{buzz}, 60.3) +- (2.4, 2.4)
    };
    \end{axis}

\end{tikzpicture}
\end{minipage}
\begin{minipage}[t]{0.31\textwidth}
    \begin{tikzpicture}[
            every node/.style={font=\fontfamily{cmr}\selectfont},
            every label/.style={font=\fontfamily{cmr}\selectfont}]
    \begin{axis}[
        ybar, 
        title = {\parbox[c]{\linewidth}{\centering \fontfamily{cmr}\selectfont AP\\Posterior mean system: $\vb{H}\bm{v}=\bm{y}$}},
        bar width=0.15cm, 
        symbolic x coords={\texttt{3drd}, \texttt{pol}, \texttt{pro}, \texttt{bike}, \texttt{buzz}}, 
        xtick=data, 
        xticklabel style={rotate=45, anchor=east, text height=1.5ex, text depth=0.25ex},
        xlabel style={yshift=-8pt},
        axis x line=bottom,   
        axis y line*=left,               
        axis line style={-},
        y axis line style={draw=none},
        ymajorgrids=true,      
        grid style={solid,gray!30}, 
        yticklabels={},
        xticklabels={},
        ylabel style={yshift=3pt},
        ymin=0, 
        ymax=100, 
        height=4cm,
        width=6.6cm,
        enlarge y limits = -10, 
        enlarge x limits=0.15, 
        legend style={at={(0.5,0.6)}, anchor=south, legend columns=-1, draw=none}, 
    ]
    \addplot+[
        blue, fill=blue!50,
        error bars/.cd,
        y dir=both, 
        y explicit 
    ] 
    coordinates {
        (\texttt{3drd}, 100) +- (0, 0)
        (\texttt{pol}, 100) +- (0, 0)
        (\texttt{pro}, 100) +- (0, 0)
        (\texttt{bike}, 100) +- (0, 0)
        (\texttt{buzz}, 100) +- (0, 0)
    };
    
    \addplot+[
        red, fill=red!50,
        error bars/.cd,
        y dir=both,
        y explicit
    ] 
    coordinates {
        (\texttt{3drd}, 41.2) +- (4.1, 4.1)
        (\texttt{pol}, 29.0) +- (3.3, 3.3)
        (\texttt{pro}, 30.5) +- (9.2, 9.2)
        (\texttt{bike}, 24.6) +- (14.9, 14.9)
        (\texttt{buzz}, 8.0) +- (4.3, 4.3)
    };
    \addplot+[
        custom_green, fill=custom_green!50,
        error bars/.cd,
        y dir=both,
        y explicit
    ] 
    coordinates {
        (\texttt{3drd}, 40.4) +- (4.6, 4.6)
        (\texttt{pol}, 28.8) +- (2.8, 2.8)
        (\texttt{pro}, 29.7) +- (8.5, 8.5)
        (\texttt{bike}, 10.5) +- (6.7, 6.7)
        (\texttt{buzz}, 7.7) +- (4.1, 4.1)
    };\addplot+[
        custom_yellow, fill=custom_yellow!50,
        error bars/.cd,
        y dir=both,
        y explicit
    ] 
    coordinates {
        (\texttt{3drd}, 41.2) +- (4.4, 4.4)
        (\texttt{pol}, 29.3) +- (2.5, 2.5)
        (\texttt{pro}, 29.9) +- (9.0, 9.0)
        (\texttt{bike}, 10.7) +- (6.7, 6.7)
        (\texttt{buzz}, 8.0) +- (4.3, 4.3)
    };
    \end{axis}
\end{tikzpicture}
\end{minipage}
\begin{minipage}[t]{0.31\textwidth}
\begin{tikzpicture}[
            every node/.style={font=\fontfamily{cmr}\selectfont},
            every label/.style={font=\fontfamily{cmr}\selectfont}]
\begin{axis}[
        ybar, 
        title = {\parbox[c]{\linewidth}{\centering \fontfamily{cmr}\selectfont CG\\ \quad}},        bar width=0.15cm, 
        symbolic x coords={\texttt{3drd}, \texttt{pol}, \texttt{pro}, \texttt{bike}, \texttt{buzz}}, 
        xtick=data, 
        xticklabel style={rotate=45, anchor=east, text height=1.5ex, text depth=0.25ex},
        xlabel style={yshift=-8pt},
        axis x line=bottom,   
        axis y line*=left,               
        axis line style={-},
        y axis line style={draw=none},
        ymajorgrids=true,      
        grid style={solid,gray!30}, 
        yticklabels={},
        xticklabels={},
        ylabel style={yshift=3pt},
        ymin=0, 
        ymax=100, 
        height=4cm,
        width=6.6cm,
        enlarge y limits = -10, 
        enlarge x limits=0.15, 
        legend style={at={(0.5,0.6)}, anchor=south, legend columns=-1, draw=none}, 
    ]
    \addplot+[
        blue, fill=blue!50,
        error bars/.cd,
        y dir=both, 
        y explicit 
    ] 
    coordinates {
        (\texttt{3drd}, 100) +- (0, 0)
        (\texttt{pol}, 100) +- (0, 0)
        (\texttt{pro}, 100) +- (0, 0)
        (\texttt{bike}, 100) +- (0, 0)
        (\texttt{buzz}, 100) +- (0, 0)
    };
    
    \addplot+[
        red, fill=red!50,
        error bars/.cd,
        y dir=both,
        y explicit
    ] 
    coordinates {
        (\texttt{3drd}, 70.2) +- (4.1, 4.1)
        (\texttt{pol}, 71.7) +- (3.2, 3.2)
        (\texttt{pro}, 75.2) +- (8.1, 8.1)
        (\texttt{bike}, 64.0) +- (3.4, 3.4)
        (\texttt{buzz}, 72.2) +- (11.7, 11.7)
    };
    \addplot+[
        custom_green, fill=custom_green!50,
        error bars/.cd,
        y dir=both,
        y explicit
    ] 
    coordinates {
        (\texttt{3drd}, 69.3) +- (4.1, 4.1)
        (\texttt{pol}, 67.5) +- (2.7, 2.7)
        (\texttt{pro}, 76.2) +- (8.6, 8.6)
        (\texttt{bike}, 82.6) +- (7.4, 7.4)
        (\texttt{buzz}, 77.9) +- (7.2, 7.2)
    };\addplot+[
        custom_yellow, fill=custom_yellow!50,
        error bars/.cd,
        y dir=both,
        y explicit
    ] 
    coordinates {
        (\texttt{3drd}, 67.5) +- (3.1, 3.1)
        (\texttt{pol}, 66.9) +- (1.8, 1.8)
        (\texttt{pro}, 73.4) +- (8.1, 8.1)
        (\texttt{bike}, 64.1) +- (3.8, 3.8)
        (\texttt{buzz}, 67.2) +- (7.0, 7.0)
    };
    \end{axis}
\end{tikzpicture}
\end{minipage}

\begin{center}
    \vspace{-0.3cm}
\hspace{0.5cm}\fontfamily{cmr}\selectfont \quad Posterior sample system: \bf $\vb{H}\bm{v} = f(\vb{X} )+\vb*{\epsilon}$
\end{center}
\vspace{-5pt}
\begin{minipage}[t]{0.345\textwidth}
    \begin{tikzpicture}[
            every node/.style={font=\fontfamily{cmr}\selectfont},
            every label/.style={font=\fontfamily{cmr}\selectfont}]
    \begin{axis}[
        ybar, 
        bar width=0.15cm, 
        symbolic x coords={\texttt{3drd}, \texttt{pol}, \texttt{pro}, \texttt{bike}, \texttt{buzz}}, 
        xtick=data, 
        xticklabel style={rotate=45, anchor=east, text height=1.5ex, text depth=0.25ex},
        xlabel style={yshift=-8pt},
        axis x line=bottom,   
        axis y line*=left,               
        axis line style={-},
        ymajorgrids=true,      
        grid style={solid,gray!30}, 
        yticklabel={\fontfamily{cmr}\selectfont\pgfmathprintnumber{\tick}},
        ylabel style={yshift=-8pt},
        ymin=0, 
        ymax=100, 
        height=4cm,
        width=6.6cm,
        enlarge y limits = -10, 
        enlarge x limits=0.15, 
        legend style={at={(0.5,0.6)}, anchor=south, legend columns=-1, draw=none}, 
    ]
    \addplot+[
        blue, fill=blue!50,
        error bars/.cd,
        y dir=both, 
        y explicit 
    ] 
    coordinates {
        (\texttt{3drd}, 100) +- (0, 0)
        (\texttt{pol}, 100) +- (0, 0)
        (\texttt{pro}, 100) +- (0, 0)
        (\texttt{bike}, 100) +- (0, 0)
        (\texttt{buzz}, 100) +- (0, 0)
    };
    
    \addplot+[
        red, fill=red!50,
        error bars/.cd,
        y dir=both,
        y explicit
    ] 
    coordinates {
        (\texttt{3drd}, 71.6) +- (5.7, 5.7)
        (\texttt{pol}, 64.7) +- (4.7, 4.7)
        (\texttt{pro}, 71.0) +- (15.7, 15.7)
        (\texttt{bike}, 53.1) +- (7.9, 7.9)
        (\texttt{buzz}, 62.1) +- (3.6, 3.6)
    };
    \addplot+[
        custom_green, fill=custom_green!50,
        error bars/.cd,
        y dir=both,
        y explicit
    ] 
    coordinates {
        (\texttt{3drd}, 63.9) +- (2.0, 2.0)
        (\texttt{pol}, 61.4) +- (4.5, 4.5)
        (\texttt{pro}, 66.6) +- (9.8, 9.8)
        (\texttt{bike}, 52.8) +- (7.1, 7.1)
        (\texttt{buzz}, 61.7) +- (5.1, 5.1)
    };\addplot+[
        custom_yellow, fill=custom_yellow!50,
        error bars/.cd,
        y dir=both,
        y explicit
    ] 
    coordinates {
        (\texttt{3drd}, 62.8) +- (3.4, 3.4)
        (\texttt{pol}, 59.2) +- (7.5, 7.5)
        (\texttt{pro}, 60.9) +- (9.5, 9.5)
        (\texttt{bike}, 46.1) +- (7.7, 7.7)
        (\texttt{buzz}, 55.7) +- (4.9, 4.9)
    };
    \end{axis}
\end{tikzpicture}
\end{minipage}
\begin{minipage}[t]{0.31\textwidth}
    \begin{tikzpicture}[
            every node/.style={font=\fontfamily{cmr}\selectfont},
            every label/.style={font=\fontfamily{cmr}\selectfont}]
    \begin{axis}[
        ybar, 
        bar width=0.15cm, 
        symbolic x coords={\texttt{3drd}, \texttt{pol}, \texttt{pro}, \texttt{bike}, \texttt{buzz}}, 
        xtick=data, 
        xticklabel style={rotate=45, anchor=east, text height=1.5ex, text depth=0.25ex},
        xlabel style={yshift=-8pt},
        axis x line=bottom,   
        axis y line*=left,               
        axis line style={-},
        y axis line style={draw=none},
        ymajorgrids=true,      
        grid style={solid,gray!30}, 
        yticklabels={},
        ylabel style={yshift=3pt},
        ymin=0, 
        ymax=100, 
        height=4cm,
        width=6.6cm,
        enlarge y limits = -10, 
        enlarge x limits=0.15, 
        legend style={at={(0.5,0.6)}, anchor=south, legend columns=-1, draw=none}, 
    ]
    \addplot+[
        blue, fill=blue!50,
        error bars/.cd,
        y dir=both, 
        y explicit 
    ] 
    coordinates {
        (\texttt{3drd}, 100) +- (0, 0)
        (\texttt{pol}, 100) +- (0, 0)
        (\texttt{pro}, 100) +- (0, 0)
        (\texttt{bike}, 100) +- (0, 0)
        (\texttt{buzz}, 100) +- (0, 0)
    };
    
    \addplot+[
        red, fill=red!50,
        error bars/.cd,
        y dir=both,
        y explicit
    ] 
    coordinates {
        (\texttt{3drd}, 42.8) +- (3.8, 3.8)
        (\texttt{pol}, 35.9) +- (5.0, 5.0)
        (\texttt{pro}, 28.3) +- (10.8, 10.8)
        (\texttt{bike}, 5.8) +- (5.2, 5.2)
        (\texttt{buzz}, 5.4) +- (1.7, 1.7)
    };
    \addplot+[
        custom_green, fill=custom_green!50,
        error bars/.cd,
        y dir=both,
        y explicit
    ] 
    coordinates {
        (\texttt{3drd}, 42.2) +- (4.0, 4.0)
        (\texttt{pol}, 35.8) +- (5.1, 5.1)
        (\texttt{pro}, 29.8) +- (11.7, 11.7)
        (\texttt{bike}, 5.7) +- (5.1, 5.1)
        (\texttt{buzz}, 5.2) +- (1.8, 1.8)
    };\addplot+[
        custom_yellow, fill=custom_yellow!50,
        error bars/.cd,
        y dir=both,
        y explicit
    ] 
    coordinates {
        (\texttt{3drd}, 41.7) +- (4.5, 4.5)
        (\texttt{pol}, 35.9) +- (5.0, 5.0)
        (\texttt{pro}, 28.1) +- (10.7, 10.7)
        (\texttt{bike}, 5.6) +- (5.1, 5.1)
        (\texttt{buzz}, 5.4) +- (1.7, 1.7)
    };
    \end{axis}

\end{tikzpicture}
\end{minipage}
\begin{minipage}[t]{0.31\textwidth}
\begin{tikzpicture}[
            every node/.style={font=\fontfamily{cmr}\selectfont},
            every label/.style={font=\fontfamily{cmr}\selectfont}]
\begin{axis}[
        ybar, 
        bar width=0.15cm, 
        symbolic x coords={\texttt{3drd}, \texttt{pol}, \texttt{pro}, \texttt{bike}, \texttt{buzz}}, 
        xtick=data, 
        xticklabel style={rotate=45, anchor=east, text height=1.5ex, text depth=0.25ex},
        xlabel style={yshift=-8pt},
        axis x line=bottom,   
        axis y line*=left,               
        axis line style={-},
        y axis line style={draw=none},
        ymajorgrids=true,      
        grid style={solid,gray!30}, 
        yticklabels={},
        ylabel style={yshift=3pt},
        ymin=0, 
        ymax=100, 
        height=4cm,
        width=6.6cm,
        enlarge y limits = -10, 
        enlarge x limits=0.15, 
        legend style={at={(0.5,0.6)}, anchor=south, legend columns=-1, draw=none}, 
    ]
    \addplot+[
        blue, fill=blue!50,
        error bars/.cd,
        y dir=both, 
        y explicit 
    ] 
    coordinates {
        (\texttt{3drd}, 100) +- (0, 0)
        (\texttt{pol}, 100) +- (0, 0)
        (\texttt{pro}, 100) +- (0, 0)
        (\texttt{bike}, 100) +- (0, 0)
        (\texttt{buzz}, 100) +- (0, 0)
    };
    
    \addplot+[
        red, fill=red!50,
        error bars/.cd,
        y dir=both,
        y explicit
    ] 
    coordinates {
        (\texttt{3drd}, 72.5) +- (3.0, 3.0)
        (\texttt{pol}, 75.7) +- (4.0, 4.0)
        (\texttt{pro}, 70.8) +- (11.3, 11.3)
        (\texttt{bike}, 54.1) +- (5.5, 5.5)
        (\texttt{buzz}, 73.4) +- (7.5, 7.5)
    };
    \addplot+[
        custom_green, fill=custom_green!50,
        error bars/.cd,
        y dir=both,
        y explicit
    ] 
    coordinates {
        (\texttt{3drd}, 70.5) +- (3.5, 3.5)
        (\texttt{pol}, 72.7) +- (3.9, 3.9)
        (\texttt{pro}, 77.3) +- (9.7, 9.7)
        (\texttt{bike}, 76.9) +- (10.3, 10.3)
        (\texttt{buzz}, 79.5) +- (6.9, 6.9)
    };\addplot+[
        custom_yellow, fill=custom_yellow!50,
        error bars/.cd,
        y dir=both,
        y explicit
    ] 
    coordinates {
        (\texttt{3drd}, 68.4) +- (6.3, 6.3)
        (\texttt{pol}, 71.1) +- (2.3, 2.3)
        (\texttt{pro}, 68.9) +- (12.0, 12.0)
        (\texttt{bike}, 54.1) +- (6.9, 6.9)
        (\texttt{buzz}, 67.3) +- (6.5, 6.5)
    };
    \end{axis}
\end{tikzpicture}
\end{minipage}
\end{minipage}

\begin{center}
    \vspace{-0.3cm}
\fontfamily{cmr}\selectfont \hspace{1cm} Dataset
\end{center}
    \vspace{-0.2cm}
 \hspace{1cm} \input{legend}

\caption{Number of solver iterations required for convergence with the warm-start methods shown as a percentage of the iterations required with cold-starting, with mean and standard deviation across trials.}\label{fig:solver_iterations_samples}
\end{figure}

%% file: Appendices/bayes_opt_results.tex


\begin{figure}[H]
    \centering
    \includegraphics[width=0.9\linewidth]{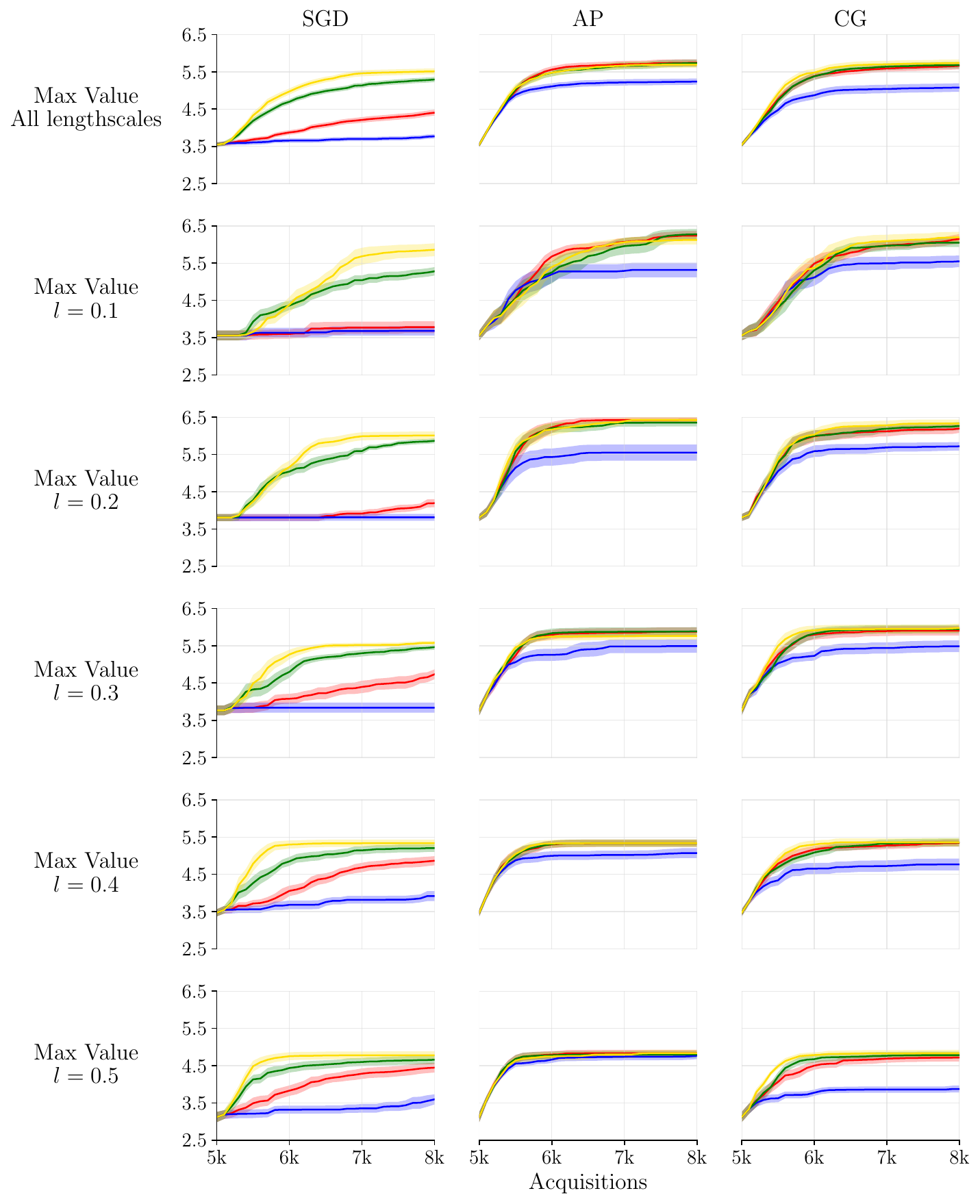}
    \vspace{-10pt}
     \input{legend}
    \caption{Maximum value found on the objective function, as a function of data point acquisitions.}
    \label{fig:max_fn_all_lengthscales}
\end{figure}

\begin{figure}
    \centering
    \includegraphics[width=0.9\linewidth]{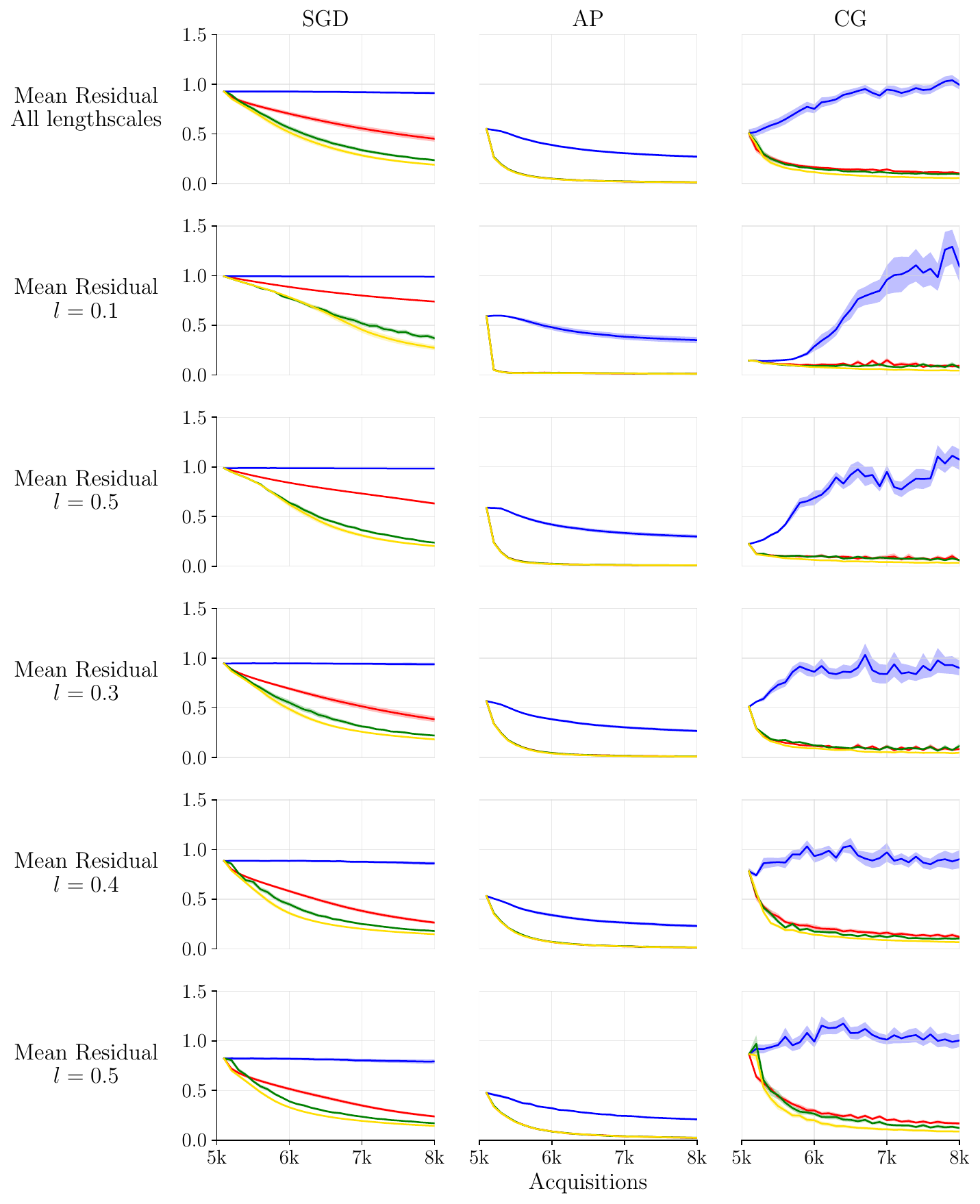}
    \vspace{-10pt}
     \input{legend}
    \caption{Final residual of the posterior mean solve, as a function of data point acquisitions.}
    \label{fig:mean_residual_all_lengthscales}
\end{figure}

\begin{figure}
    \centering
    \includegraphics[width=0.9\linewidth]{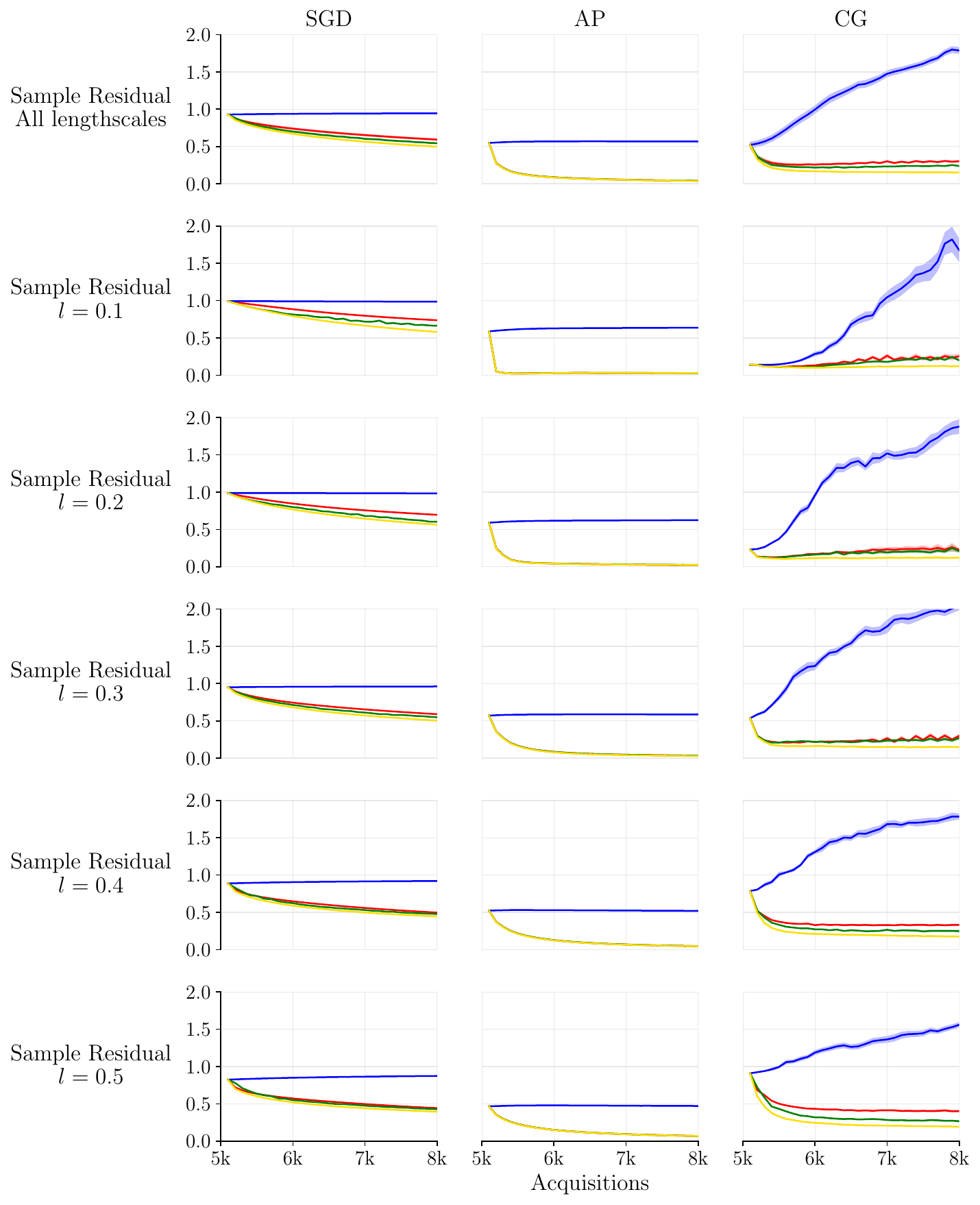}
    \vspace{-10pt}
      \input{legend}
    \caption{Average final residual of the 100 posterior sample solves, as a function of data point acquisitions.}
    \label{fig:sample_residual_all_lengthscales}
\end{figure}

%% file: Appendices/Thompson_method.tex
We performed the following steps to find the input locations which maximize each posterior sample, following the method used in \citet{sampling_from_gp_posteriors}, at one-tenth of the scale due to hardware limitations:

\begin{enumerate}
    \item Propose 5000 new input locations. \citet{sampling_from_gp_posteriors} suggests a mix of random sampling strategies to efficiently explore the feature space, locate the maximum, and capture features at the Nyquist frequency. Using their strategy, 10\% of the new locations are exploration points uniformly sampled in the input space. The remaining 90\% are sampled from $\mathcal{N}\left(0,\left(\frac{l}{2}\right)^2\right)$ distributions
    centered on the known data points, with each center selected with probability proportional to its value. This exploits known points so that a large proportion of newly proposed locations are located around the higher scoring known points on the objective function.

    \item Each of the 100 GP posterior samples is evaluated at these 5000 input locations. The highest point on each posterior sample is recorded. This process of proposing 5000 new locations and selecting the highest on each posterior sample is repeated 30 times, to obtain 30 highest points for each posterior sample.

    \item Perform gradient ascent on the 30 points on each posterior sample using Adam. After 100
    Adam steps, select the highest point on each posterior sample after gradient ascent, which is expected to be close to the maximum of each posterior sample.

    \item The point which maximizes each posterior sample is used as a location to evaluate the objective function.
\end{enumerate}

\begin{figure}[hb]
    \centering
    \includegraphics[width=1.0\linewidth]{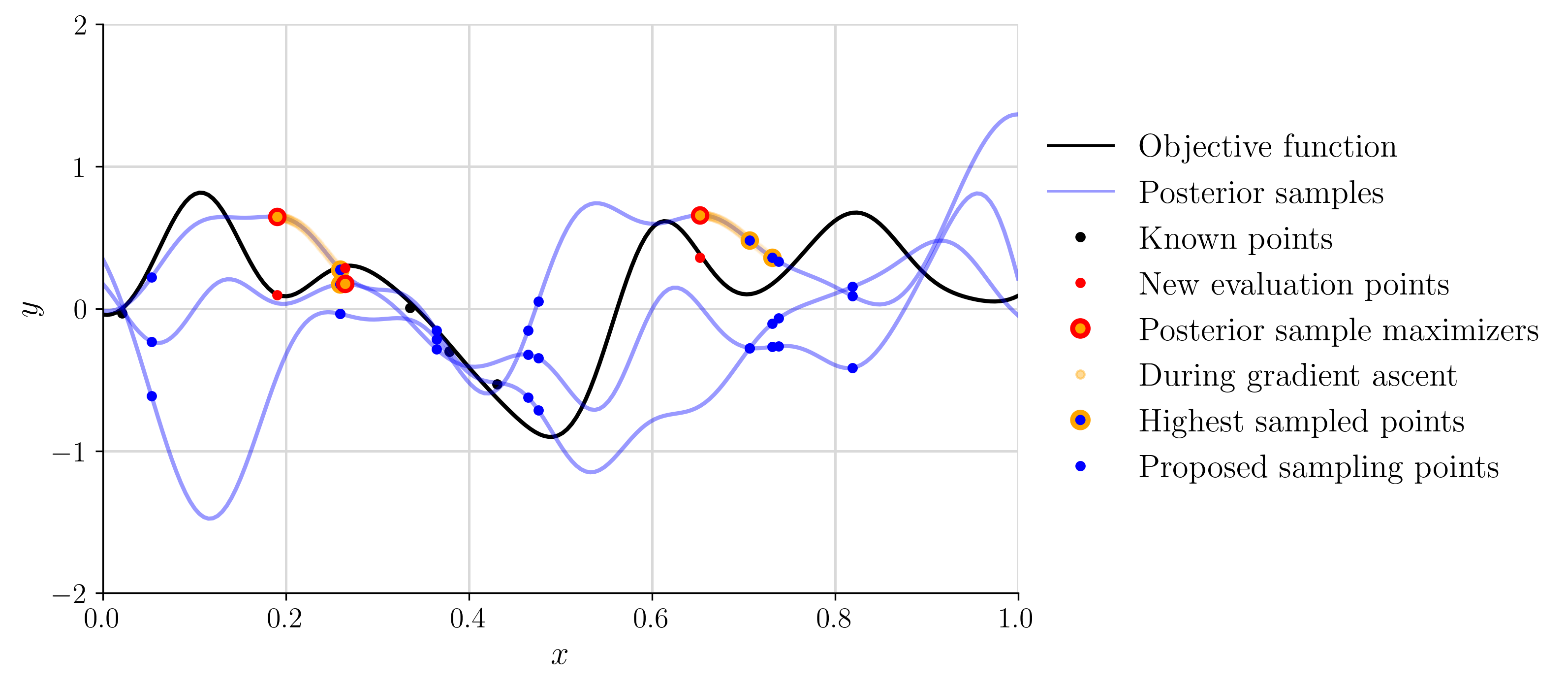}
    \caption{Posterior sample maximizing step of parallel Thompson sampling for 1D Bayesian optimization}
    \label{fig:1dillustration}
\end{figure}

\Cref{fig:1dillustration} illustrates this procedure during a single stage of parallel Thompson sampling on a 1D optimization problem.

